\documentclass[journal,onecolumn]{IEEEtran}

\IEEEoverridecommandlockouts                              

\usepackage{multirow}
\usepackage{graphicx}
\usepackage{physics}
\usepackage{amsmath}
\usepackage{cite}
\usepackage{amsmath}
\usepackage{algorithm}
\usepackage{algorithmic}
\usepackage{graphicx}
\usepackage{multirow}
\usepackage{booktabs}
\usepackage{amssymb}
\usepackage[hidelinks]{hyperref}
\usepackage{cleveref}
\usepackage{changepage}\usepackage{xcolor}  
\usepackage{longtable} 
\usepackage{array} 
\usepackage{float}
\usepackage{placeins}
\usepackage{afterpage}
\usepackage{lipsum}
\DeclareGraphicsExtensions{.pdf,.jpeg,.png}

\Crefname{figure}{Fig.}{Figs.}

\title{\LARGE \bf SAM: Semi-Active Mechanism for Extensible Continuum Manipulator and Real-time Hysteresis Compensation Control Algorithm}

\author{Junhyun Park$^{1*}$, Seonghyeok Jang$^{1*}$, Myeongbo Park$^1$, Hyojae Park$^1$, Jeonghyeon Yoon$^1$, and Minho Hwang$^{1\dagger}$}

\begin{document}

\maketitle
\thispagestyle{empty}
\pagestyle{empty}
\vspace{-0.8cm}
\par
$^1$ Department of Robotics and Mechatronics Engineering, Daegu Gyeongbuk Institute of Science and Technology (DGIST), Daegu 42988, Republic of Korea
{\tt\footnotesize \{sean05071, jshtop1, qkraudqh23, hyojae, yjh1434\}@dgist.ac.kr, gkgk1215@gmail.com }\par
\vspace{0.5cm}
{\footnotesize $^*$ These authors contributed equally to this work.\par}
{\footnotesize $^\dagger$ Corresponding author}

\vspace{0.5cm}
\begin{abstract}
\textbf{\normalsize Background:}

\begin{adjustwidth}{3em}{3em}
Cable-Driven Continuum Manipulators (CDCMs) enable scar-free procedures but face limitations in workspace and control accuracy due to hysteresis.
\end{adjustwidth}

\textbf{\normalsize Methods:}

\begin{adjustwidth}{3em}{3em}
We introduce an extensible CDCM with a Semi-active Mechanism (SAM) and develop a real-time hysteresis compensation control algorithm using a Temporal Convolutional Network (TCN) based on data collected from fiducial markers and RGBD sensing.
\end{adjustwidth}

\textbf{\normalsize Results:}

\begin{adjustwidth}{3em}{3em}
Performance validation shows the proposed controller significantly reduces hysteresis by up to 69.5\% in random trajectory tracking test and approximately 26\% in the box pointing task.
\end{adjustwidth}

\textbf{\normalsize Conclusion:}

\begin{adjustwidth}{3em}{3em}
The SAM mechanism enables access to various lesions without damaging surrounding tissues. The proposed controller with TCN-based compensation effectively predicts hysteresis behavior and minimizes position and joint angle errors in real-time, which has the potential to enhance surgical task performance. 
\end{adjustwidth}
\begin{flushleft}
\end{flushleft}
\end{abstract}

\section{Introduction}

Rigid surgical manipulator encounter difficulties in accessing lesions, especially in surgeries involving internal organs like the small and large intestines \cite{Surgical_robot, review3_1, c2, optimizingbase}. In contrast, Cable-Driven Continuum Manipulators (CDCMs), with their flexible and bendable property, are expected to enable minimally invasive surgery by navigating through complex internal organs \cite{K-Flex,TROS, MASTER, ESD, review3_5, strong_continuum, hwnag_robotic_2020}. CDCMs are emerging as a next-generation surgical manipulation technology.

Miniaturization of CDCMs for endoscopic surgery arise challenge due to size constraints \cite{daVeiga2020ChallengesOC}. While miniaturization is crucial for safe insertion without damaging tissue, it inherently restricts the workspace \cite{mini_cont, ContinuumHardware}, making it difficult for surgeons to navigate and access diverse target lesions. This conflict between miniaturization and workspace necessitates the development of novel CDCM \cite{mechansim}. Additionally, endoscopic CDCMs leverage cable actuation for insertion into the body, with motors positioned outside the patient. However, various factors such as cable elongation\cite{extension}, friction\cite{friction}, twist\cite{twist}, and coupling\cite{coupling} induce enlarging hysteresis. This hysteresis behavior hinders precise control of movement, impacting surgical accuracy and potentially extending operating time \cite{kimhansol}. These limitations pose a risk to the broader adoption of CDCMs in real-world surgical settings \cite{Challenge}.

\begin{figure}[t!]
  \centering
  \includegraphics[width=0.55\linewidth]{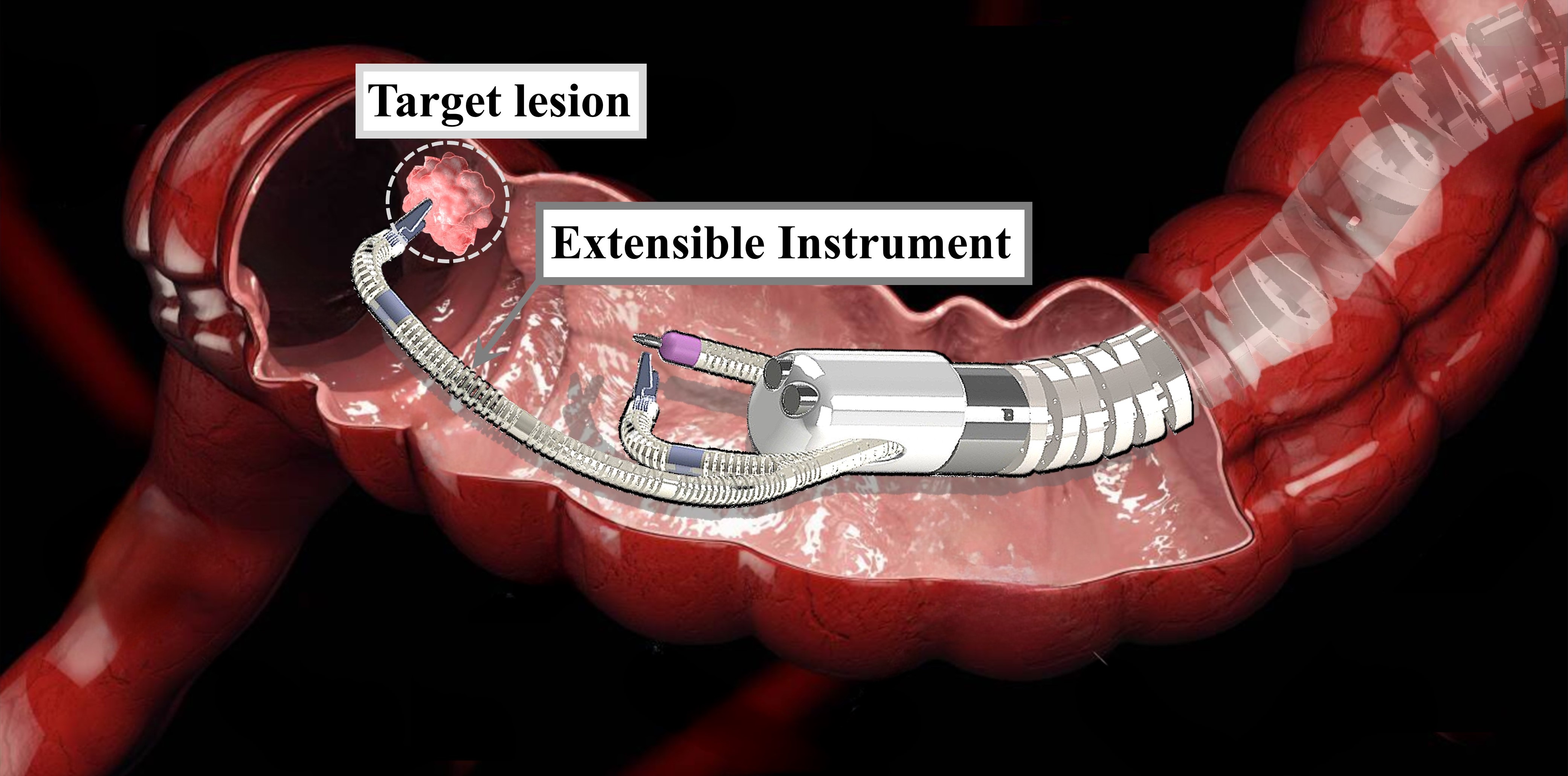}
  \caption{\textbf{Concept Design of the Proposed Mechanism:} By utilizing the Semi-active segment, the SAM manipulator can access to the target lesions.}
  \label{conceptual}
\end{figure}

\begin{figure}[t!]
  \centering
  \includegraphics[width=0.6\linewidth]{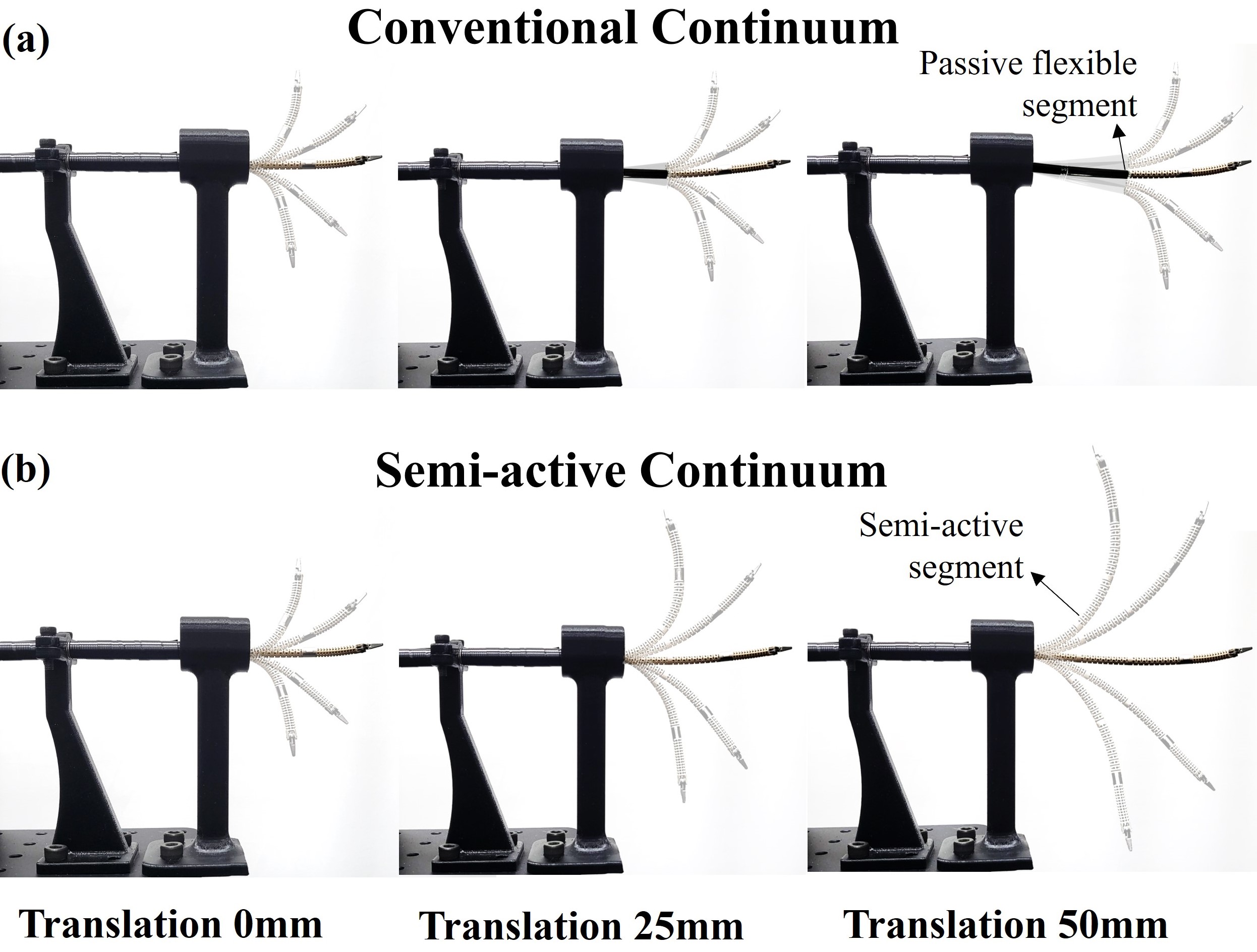}
  \caption{\textbf {Comparison of Workspace during Translation between Conventional Continuum and Semi-active Continuum}: (a) Conventional CDCM with fixed-length segments and a passive flexible segment. This design limits the workspace regardless of manipulator translation and restricts accessibility to the target region. (b) The proposed manipulator with a Semi-active segment that lengthens during {translation through the driving part}, resulting in an increased workspace within its operational space.}
  \label{demo_figure}
\end{figure}

\begin{figure}[t!]
  \centering
  \includegraphics[width=0.95\linewidth]{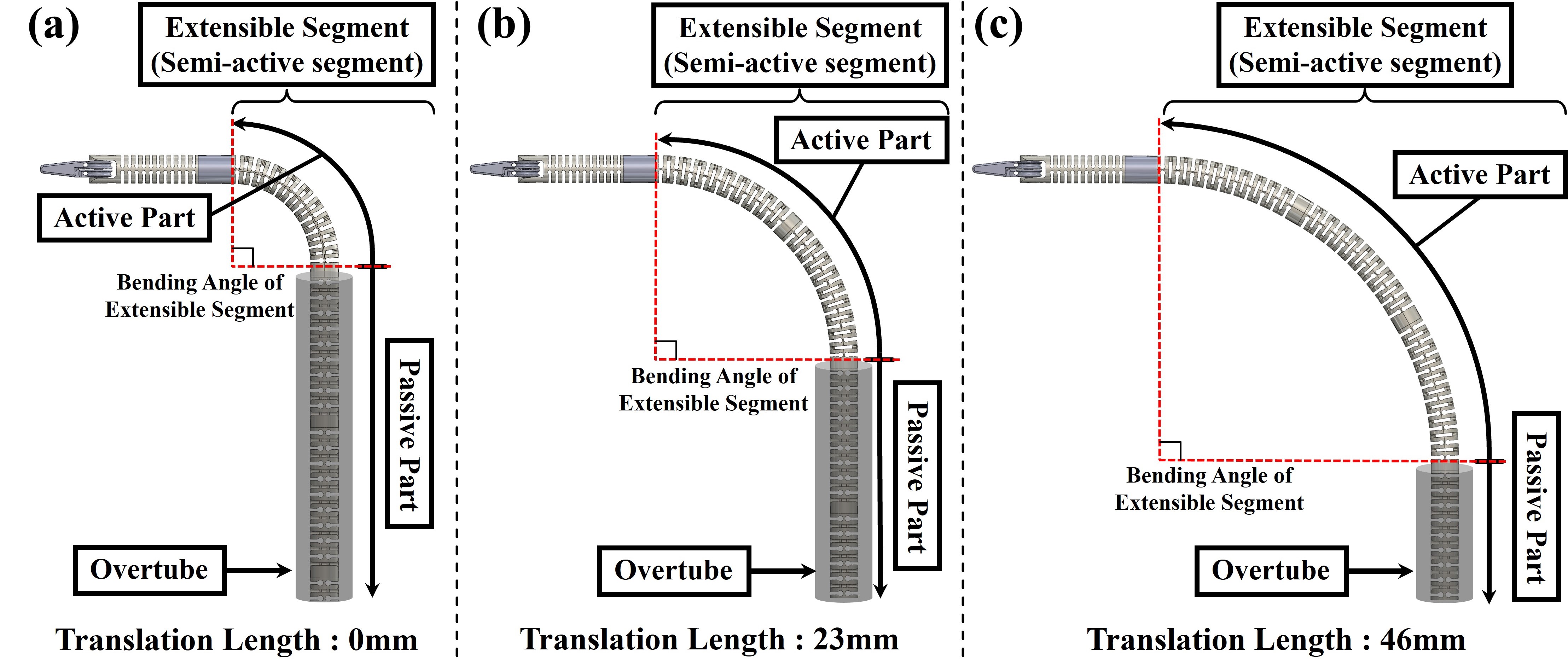}
  \caption{\textbf{Extension Principles and Features of the Semi-active Segment :} {The Semi-active segment integrates both active and passive parts. The active part is directly driven to enable bending motion, while the passive part is flexible and transmits the driving force. As the translation length increases, the active part extends. A key feature of the Semi-active segment is its ability to maintain a consistent bending angle during this extension. The segment maintains a consistent bending angle of 90 degrees during extension: (a) At a translation length of 0 mm. (b) At a translation length of 23 mm. (c) At a translation length of 46 mm.}}
  \label{Semi-active Mechansim}
\end{figure}

This study introduces the Semi-active Mechanism (SAM)  for compact extensible endoscopic CDCMs. Compared to conventional continuum manipulators, the proposed mechanism features ensures that the workspace proportionally expands as the instrument undergoes axial translation (refer to \Cref{conceptual}). {In endoscopic CDCMs, the actively driven parts are referred to as active segments, while parts that transmit motion but do not move directly are termed passive flexible segments (e.g., insertion tube).} In conventional CDCMs, the proximal segment is directly connected to a passive flexible segment, resulting in a fixed workspace regardless of translation. By comparison, the proposed SAM functions as an active segment when protruded, extending the workspace, and when retracted, the portion that enters the overtube behaves like a passive flexible segment (refer to \Cref{demo_figure} and \Cref{Semi-active Mechansim}).

The proposed extension mechanism enhances the workspace, but introduces the challenge of hysteresis model variation as the Semi-active segment length increases. To address this, we propose a real-time hysteresis compensation control algorithm for the extensible continuum manipulator. The approach involves constructing a hysteresis dataset using an RGBD camera and fiducial markers to capture the relationship between command joint angles and the physical joint angles. We then employ a Temporal Convolutional Network (TCN) to model the complex hysteresis behavior. This trained TCN model estimates the command joint angles for the inputted physical joint angles. Based on this estimation, we develop a real-time control algorithm that actively compensates for hysteresis, leading to enhanced control accuracy. Finally, the performance of the TCN-based compensation method is validated through random joint trajectory tracking and box pointing tasks, demonstrating significant hysteresis reduction in both operational space and joint space.

The main contributions of this paper are summarized as follows: (1) Design and kinematics analysis of the proposed extensible surgical instrument, SAM, {which operates without the need for additional mechanical elements (e.g., springs, magnets, rack and pinion components), relying solely on translational motion.} (2) Proposal of real-time hysteresis compensation control algorithm in 1ms latency on the proposed instrument. (3) Validation of the proposed control algorithm with random trajectory tracking test and box pointing task suggesting significant hysteresis reduction on both operational space and joint space.

\section{Related Works}
\label{sec:related_works}
Extensible continuum manipulators have been the subject of various research efforts. These include the use of permanent magnets attached to backbone disks \cite{magnet_ex1}, incorporating springs into segments of continuum manipulators to enable extension \cite{spring_ex}, adjusting length by combining a rack and pinion structure on continuum manipulators \cite{gear_ex}, automatically attaching and detaching spacer disks on backbone-type continuum manipulators to enable extension \cite{assem_ex}, and extending the three backbones of continuum manipulators through telescopic principles \cite{tele_ex}. Most studies achieve extension and contraction through cable actuation, often incorporating additional mechanical elements such as magnets, springs, and gears. However, these additions can introduce new challenges. For instance, magnetic components often cause control instability due to electromagnetic interference in clinical settings, while gears or springs impede miniaturization efforts. {In contrast, the proposed Semi-active Mechanism (SAM) enables workspace extensibility without the need for additional mechanical components (refer to \Cref{demo_figure}-(b) and \Cref{Semi-active Mechansim}). The SAM can be implemented by sequentially attaching identical segments. Specifically, the extension principle allows the length of the active segment to increase through the translational motion of the robot. The translation of the robot can be executed through the driving part's rack and pinion (refer to \Cref{Testbed_Figure}). Unlike traditional methods using springs or gears, this design simplifies the mechanism by avoiding additional mechanical elements, making it applicable to manipulators with a diameter smaller than 5mm.}

Cable-driven surgical-assistant robots \cite{dvrk2014} and CDCMs struggles with hysteresis caused by factors like elongation \cite{extension}, friction \cite{friction}, twist \cite{twist}, and coupling \cite{coupling}. These factors hinder precise control and extend surgical completion times \cite{kimhansol}. Previous research has addressed this through analytical modeling \cite{analytic1,analytic2,analytic3,analytic4,analytic5}, learning-based methods \cite{learning_base, Hwang_Dvrk, hwangautomation2023, park2024hysteresis}, and hybrid approaches combining both techniques \cite{Kim_TSM}. Park et al. \cite{park2024hysteresis} utilized RGBD cameras and fiducial markers to estimate CDCM poses, employing a TCN-based hysteresis compensation control algorithm that achieved a 60\% reduction in hysteresis. However, their algorithm depends on various parameters and repeatedly utilizes trained models multiple time, resulting in a time latency of approximately 0.05 seconds, which is not be suitable for real-time applications.

Current hysteresis compensation research primarily focuses on general CDCM designs, which are not directly applicable to the proposed manipulator due to the significant impact of extension length on hysteresis behavior. To address this problem, we propose a novel real-time hysteresis compensation model specifically designed for the SAM. This model leverages a dataset of hysteresis behavior under various extension lengths. By utilizing a single trained model that estimates command joint angles from physical joint angles, we achieve real-time compensation control without requiring additional control parameters.

\section{Kinematics Analysis of Extensible Continuum Manipulator}
\begin{figure}[t!]
  \centering
  \includegraphics[width=0.55\linewidth]{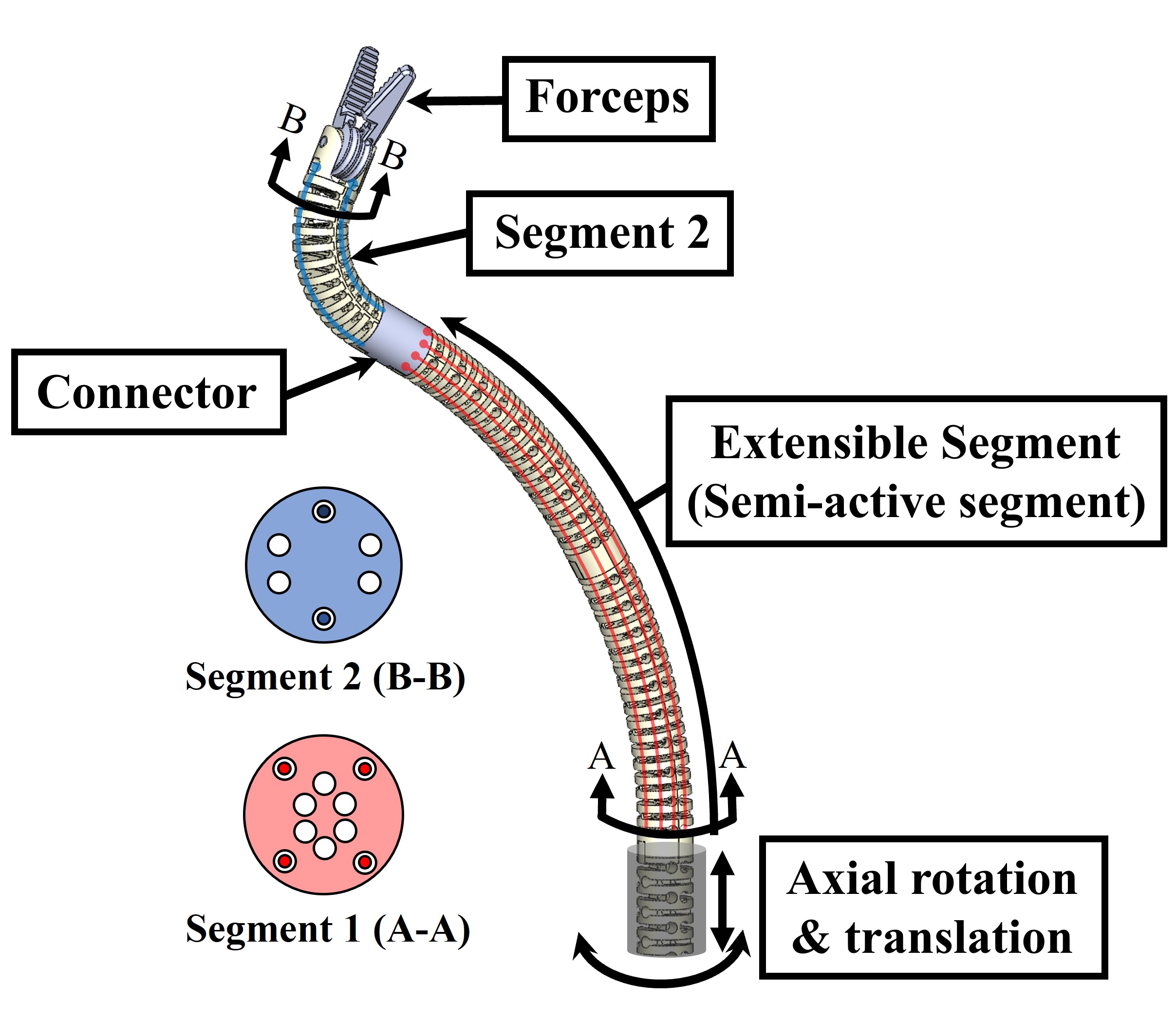}
  \caption{\textbf{Components of the Proposed Continuum Manipulator :} {Overall structure showing the main components including extensible segment (Semi-active segment), segment 2, forceps, and the axial DOFs of the base.}}
  \label{segment_Figure}
\end{figure}

{The proposed continuum manipulator comprises three main components: an extensible segment (Semi-active segment), segment 2, and forceps, as illustrated in \Cref{segment_Figure}. The joints in the extensible segment and segment 2 of the proposed continuum manipulator share identical design parameters with the flexure hinge module presented in our previous work \cite{park2024hysteresis} (e.g., module radius: 2.4 mm, disk thickness: 0.7 mm). The extensible segment is constructed by repetitively connecting the proximal bending segments, which proposed in our earlier research, to form the Semi-active segment (see \Cref{Cable_Equation}-(a)). This repeated structure enables segment extension during the robot's translational motion while maintaining a consistent angle during bending motion, which leads wider workspace compared to directly connecting proximal segments to a passive flexible part (e.g., a medical insertion tube) (refer to \Cref{Semi-active Mechansim}).}

The manipulator features 7 Degrees-of-Freedom (DOFs), including axial translation and rotation, pitch and yaw bending of {extensible} segment, pitch bending of segment 2, and yaw rotation and grasping of the forceps (\Cref{Cable_Equation}-(a)). It utilizes a total of 10 actuation cables: 4 for the extensible segment, 2 for segment 2, and 4 for the forceps.

\subsection{Forward Kinematics of Extensible Segment}
We denote the parameters of the extensible segment as $\chi_{\mu}$. As illustrated in \Cref{kinematics_Figure}-(a), these parameters are represented by $\chi_{\mu}=\left [ \kappa ,\varphi ,\vartheta  \right ]$. Specific descriptions of the coordinate frames from the manipulator's base to the end-effector (EE) are mentioned in \Cref{kinematics_Figure}-(b), and {Appendix \Cref{appendix:term}}. The kinematics of the extensible segment varies based on the protrusion of the Semi-active segment, resulting in changes to the arc length corresponding to the translation length $q_{1}$. {The arc length of the extensible segment is given by:}
\begin{gather}
    s = q_{1} + l_{1}  
\end{gather}

The curvature $\kappa$ of the segment can be represented using the components of curvature along the x-axis $\kappa _{x}$ and y-axis $\kappa _{y}$ of the local frame. The components of curvature, curvature, and the angle $\varphi$ to the bending plane can be expressed as follows:
\begin{gather}
    \kappa_{x} = \frac{q_{3}}{s},\;\; \kappa_{y} = \frac{q_{4}}{s} \nonumber \\
    \kappa = \sqrt{\kappa_{x}^{2}+\kappa_{y}^{2}}\nonumber \\
    \varphi = atan2(\kappa_{x},\kappa_{y})
\end{gather}
The angle $\vartheta$ at which the segment is bent, and the translation vector ${\mathbf{P}(s)}$ can be obtained as follows:
\begin{gather}
    \vartheta = \sqrt{{q_{3}}^{2}+{q_{4}}^{2}} \nonumber \\
    {\mathbf{P}(s)}=\begin{bmatrix}
        \dfrac{cos\varphi }{\kappa }(1-cos(\vartheta))\\[1ex]
        \dfrac{sin\varphi }{\kappa }(1-cos(\vartheta))\\[1ex]
        \dfrac{1}{\kappa }(sin(\vartheta))
\end{bmatrix}
\end{gather}

Exceptionally, when the bending angle $\vartheta \approx 0$, the radius of the arc becomes infinity, and the translation vector ${\mathbf{P}(s)}$ becomes:
\begin{gather}
{\mathbf{P}(s)}=\left [\:  0 \;\;\;   0 \;\;\;  l_{1} \: \right ]^{T}
\end{gather} 

Based on topology listed in \Cref{parameter_table}, the homogeneous transformation matrix from the base to the end of the extensible segment can be written as:
\begin{equation}
\begin{split}
{^{i-1}{\mathbf{T}}_{i}} = \begin{bmatrix}
 {\mathbf{R}_{z}}(\varphi)& 0 \\ 
 0 & 1
\end{bmatrix}  \begin{bmatrix}
 {\mathbf{R}_{x}}( \kappa s ) & {\mathbf{P}(s)} \\ 
 0 & 1
\end{bmatrix}  \begin{bmatrix}
 {\mathbf{R}_{z}}( -\varphi) & 0 \\ 
 0 & 1
\end{bmatrix}
\end{split}
\end{equation}

This transformation $\mathbf{^{i-1}{T}_{i}}$ represents the complete transformation from the base coordinate frame $O_{i-1}$ to the end coordinate frame $O_i$ of the continuum segment.

\subsection{Kinematics of Proposed Manipulator}
\label{sec:kinematics}
This section describes the forward and inverse kinematics of the proposed manipulator. Specific descriptions of the coordinate frames from the manipulator's base to the EE are mentioned in \Cref{kinematics_Figure}-(b), and {Appendix \Cref{appendix:term}} The transformation matrix from $O_{1_{b}}$ to $O'_{1_{b}}$ involves a rotation about the z-axis, resulting in:
\begin{figure}[t!]
  \centering
  \includegraphics[width=0.5\linewidth]{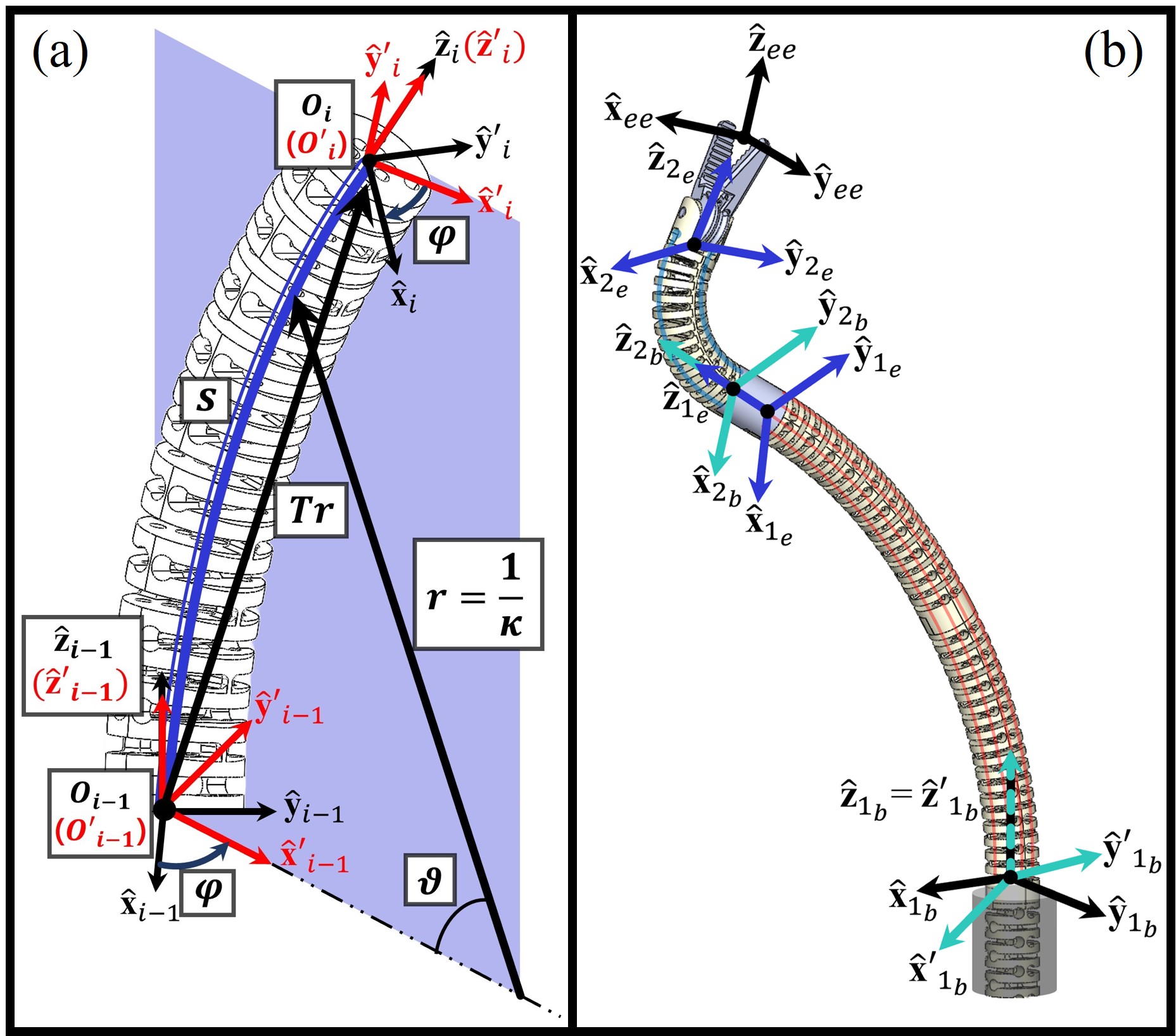}
  \caption{\textbf{Kinematics Diagram of {Continuum} Segment and the Proposed Manipulator : } (a) The coordinate frame and terminology of the continuum segment, (b) The coordinate frame and terminology of the proposed manipulator.}
  \label{kinematics_Figure}
\end{figure}
\begin{gather}
{^{1_{b}}{\mathbf{T}}_{1'_{b}}} = \begin{bmatrix}
{\mathbf{R}_{z}}(q_{2}) & 0\\ 
0 & 1
\end{bmatrix}
\end{gather} 

The transformation matrix from $O'_{1_{b}}$ to $O_{1_{e}}$ is identical to the result of the extensible segment's forward kinematics:
\begin{gather}
{^{1'_{b}}{\mathbf{T}}_{1_{e}}} \equiv \: {^{i-1}{\mathbf{T}}_{1}} 
\end{gather} 

The transformation matrix from $O_{1_{e}}$ to $O_{2_{b}}$ applies connector length along the z-axis direction. The transformation matrix from $O_{2_{b}}$ to $O_{2_{e}}$ applies the position and rotation of segment 2, bent in the pitch direction, consistent with the assumptions of the extensible segment's kinematics modeling. The expression for the transformation matrix is as follows:
\begin{gather}
{^{2_{b}}{\mathbf{T}}_{2_{e}}} = \begin{bmatrix}
{\mathbf{R}_{y}}(q_{5}) & {\mathbf{P}}(s_{2})\\ 
0 & 1
\end{bmatrix}\nonumber \\
\text{where} \;\; {\mathbf{P}}(s_{2})=\begin{bmatrix}
    \dfrac{s_{2}}{q_{5}}(1-cos(q_{5}))\\
    0 \\
    \dfrac{s_{2}}{q_{5}}(sin(q_{5}))
\end{bmatrix}
\end{gather}

The transformation matrix from $O_{2_{e}}$ to $O_{ee}$ applies yaw rotation of the forceps and length, resulting in:
\begin{gather}
{^{2_{e}}{\mathbf{T}}_{ee}} = \begin{bmatrix}
 {\mathbf{I}} & a_{3} \\ 
 0 & 1
\end{bmatrix}  \begin{bmatrix}
 {\mathbf{R}_{x}}( q_{6} ) & {^{2_{e}}\mathbf{P}_\mathrm{ee}} \\ 
 0 & 1
\end{bmatrix}  \nonumber \\
\text{where} \;\; {^{2_{e}}\mathbf{P}_{ee}}=\begin{bmatrix}
    0\\
    -sin(q_{6})d_{4} \\
    cos(q_{6})d_{4}
\end{bmatrix}
, {\mathbf{P}(s)}=\left [\:  0 \;\;\;   0 \;\;\;  l_{1} \: \right ]^{T}
\end{gather}

Finally, the transformation matrix from the manipulator's base to the EE is obtained as follows:
\begin{gather}
{^\mathrm{base}{\mathbf{T}}_\mathrm{ee}}={^{1_{b}}{\mathbf{T}}_{1'_{b}}\cdot^{1'_{b}}{\mathbf{T}}_{1_{e}}\cdot^{1_{e}}{\mathbf{T}}_{2_{b}}\cdot^{2_{b}}{\mathbf{T}}_{2_{e}}\cdot^{2_{e}}{\mathbf{T}}_\mathrm{ee}}
\end{gather} 

\begin{figure}[t!]
  \centering
  \includegraphics[width=0.90\linewidth]{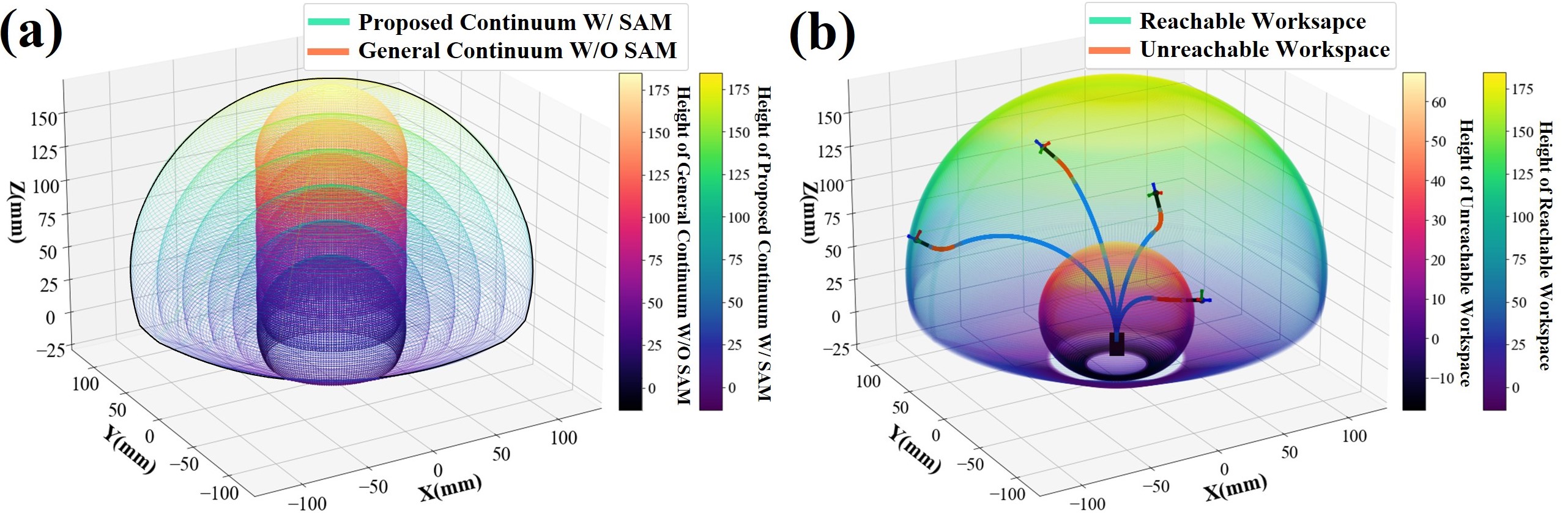}
  \caption{\textbf{Visualization of the Workspace of a Continuum Manipulator with a SAM:} (a) Difference in workspace between the general continuum manipulator W/O SAM and proposed continuum manipulator W/ SAM based on translation length, (b) The reachable and unreachable workspace of the proposed manipulator W/ SAM.}
  \label{workspace_Figure}
\end{figure}

Specific descriptions of the proposed manipulator's operational parameters are provided in {Appendix \Cref{appendix:term}} As illustrated in \Cref{workspace_Figure}-(b), we employ forward kinematics to compute the manipulator's workspace, delineating reachable and unreachable areas.

To find the solution for inverse kinematics, we apply the Broyden-Fletcher-Goldfarb-Shanno Algorithm (BFGS) to obtain a numerical solution. The equation is as follows:
\begin{gather}
{\mathbf{q}_{k+1}}={\mathbf{q}_{k}}-\alpha_{k}{\mathbf{H}_{k}}{\mathbf{J}_{k}}\nonumber\\
\underset{\mathbf{q}\in \mathbb{R}^{6}}{\text{minimize}}\:f(\mathbf{q}),\:\:\: \text{where} \: f:\mathbb{R}^{6} \rightarrow \mathbb{R}\\
f(\mathbf{q}) = \left \| {\mathbf{T}}(\mathbf{q})-{\mathbf{T}_d} \right \|
\end{gather}

Here, the subscript $k$ denotes the corresponding sequence, 
$q$ is the joint angle, $\alpha$ is the step size, ${\mathbf{J}_{k}}$ is the Jacobian matrix, ${\mathbf{H}_{k}}$ is the approximated Hessian matrix, $\mathbf{q}$ represents a vector of joint angles, including $\left \{q_1,q_2,q_3,q_4,q_5,q_6  \right \}$, and ${\mathbf{T}}(\mathbf{q})$ represents the position and orientation of Tool Center Point (TCP), and ${\mathbf{T}_d}$ represents the desired TCP position and orientation. The $\left \| \cdot  \right \|$ denotes the norm representing the magnitude of the vector.

\begin{table}[t!]
\centering
\renewcommand{\arraystretch}{1.0} 
\caption{Volume of workspace by translation, $q_{1}$}
\label{tab:workspace_table}
\begin{tabular}{c@{\hspace{0.002em}}ccccccc@{}} 
\toprule
\multirow{2}{*}{\centering\shortstack[l]{Volume\\~~~~($\times10^5\,\text{mm}^3$)}} & \multicolumn{6}{c}{$q_{1}$(mm)} \\
& 0 & 25 & 50 & 75 & 100 & 125\\ 
\midrule
\begin{tabular}[c]{@{}c@{}}Semi-active \end{tabular} & 4.56&10.76&20.24&34.05&53.0&77.9\\ 
\midrule
\begin{tabular}[c]{@{}c@{}}\ General \end{tabular} & \multicolumn{6}{@{\hspace{0.0em}}c@{\hspace{1.5em}}}{$\hspace{2.5em}4.56\hspace{2.5em}$} \\
\bottomrule
\end{tabular}
\end{table}

By minimizing $f(\mathbf{q})$, we can adjust the joint angles to move the robot to the desired target position and orientation, solving the inverse kinematics problem.

\subsection{Workspace Comparison with Conventional Continuum Manipulator} 

We calculate the workspace volumes of a continuum manipulator by using the discrete integration of Tomas Simpson method \cite{simpson_rule}. We compare the workspace volumes of the proposed manipulator and a general manipulator based on the results of forward kinematics (refer to \Cref{tab:workspace_table} and \Cref{workspace_Figure}-(a)). When the protrusion length $q_{1}$ is at its maximum of 125mm, the volume of reachable workspace of the general continuum manipulator is $456,083 \, \text{mm}^3$, whereas the volume of the reachable workspace of the proposed manipulator increase to $7,790,518 \, \text{mm}^3$. The total volume of the workspace including translation is $1,496,967\, \text{mm}^3$ for the general continuum manipulator and $7,790,518 \, \text{mm}^3$ for the proposed continuum manipulator. As shown in \Cref{workspace_Figure}-(b), the volume of the accessible workspace excluding the unreachable workspace is $7,485,460\, \text{mm}^3$ for the proposed continuum manipulator and $1,191,909\, \text{mm}^3$ for the general continuum manipulator. This indicates that the presence of SAM significantly increases the workspace volume by about 527.6\%. This implies that in future surgical applications, the proposed surgical instrument by itself can access various lesions.

\begin{figure}[t!]
  \centering
  \includegraphics[width=0.90\linewidth]{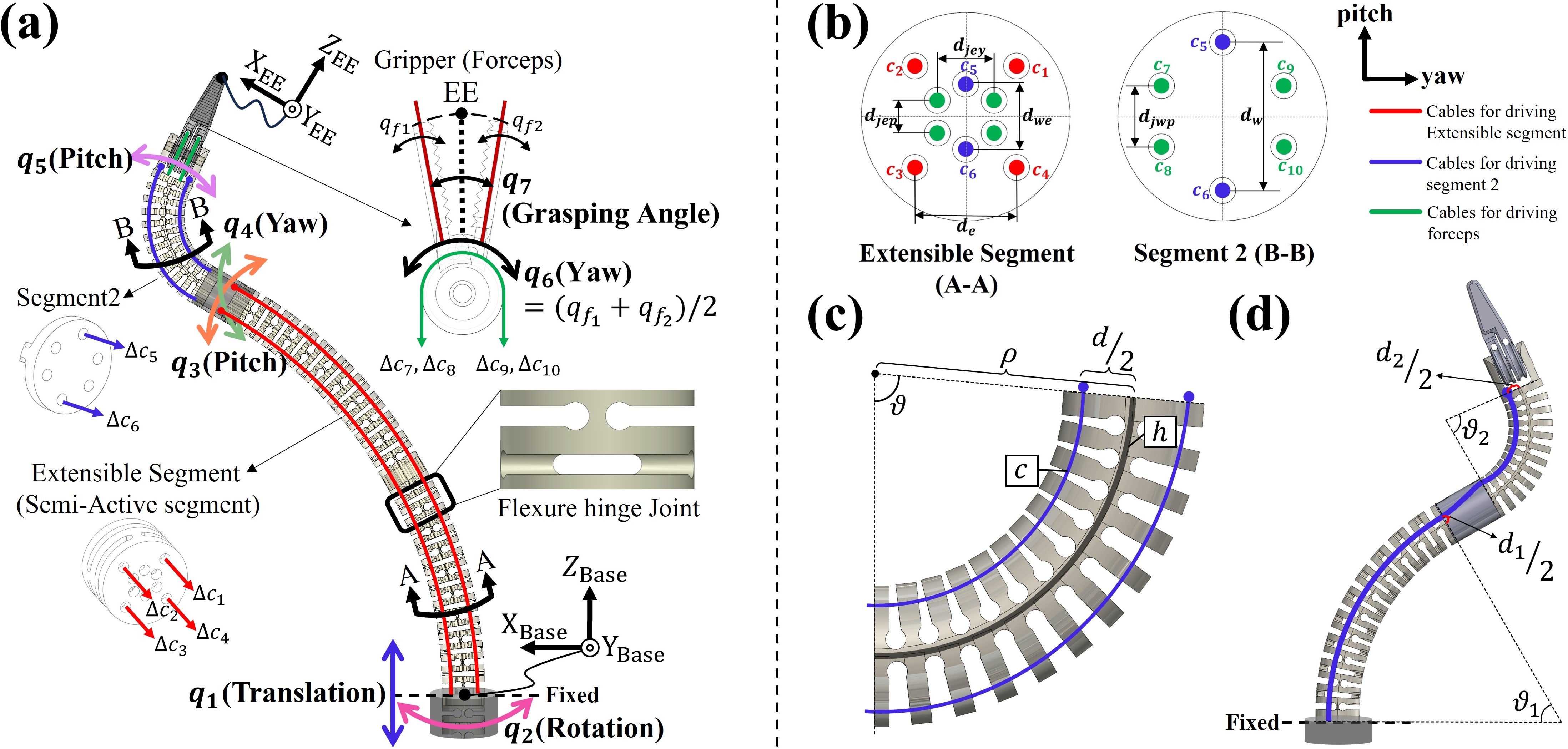}
  \caption{\textbf{{Configuration of Cables and DOFs of the Proposed Continuum Manipulator :}} {(a) Detailed illustration of the robot's cable connections and DOFs, showing the cable arrangement in the extensible segment, segment 2, and forceps, as well as the defined DOFs for the robot ($q_1$, $q_2$, $q_3$, $q_4$, $q_5$, $q_6$, $q_7$). $q_{f_1}$ and $q_{f_2}$ represent the individual DOFs for the left and right forceps, respectively, where $q_6$ can be calculated as $(q_{f_1} + q_{f_2})/2$. $q_7$, representing the grasping angle, is defined as $(q_{f_1} - q_{f_2})/2$. 
  (b) Cross-sectional views of the extensible segment (A-A) and segment 2 (B-B), showing the cable arrangements and their respective offsets within each section. (c) Geometric relationships for estimating the approximate lengths of the cables within the manipulator. 
  (d) Illustration of the coupling phenomenon of the cable in segment 2, demonstrating how the movement of segment 1 influences cable behavior in segment 2.}}
  \label{Cable_Equation}
\end{figure}

\subsection{Cable Actuation Equation}
\label{sec:control}
\begin{figure}[t!]
  \centering
  \includegraphics[width=0.55\linewidth]{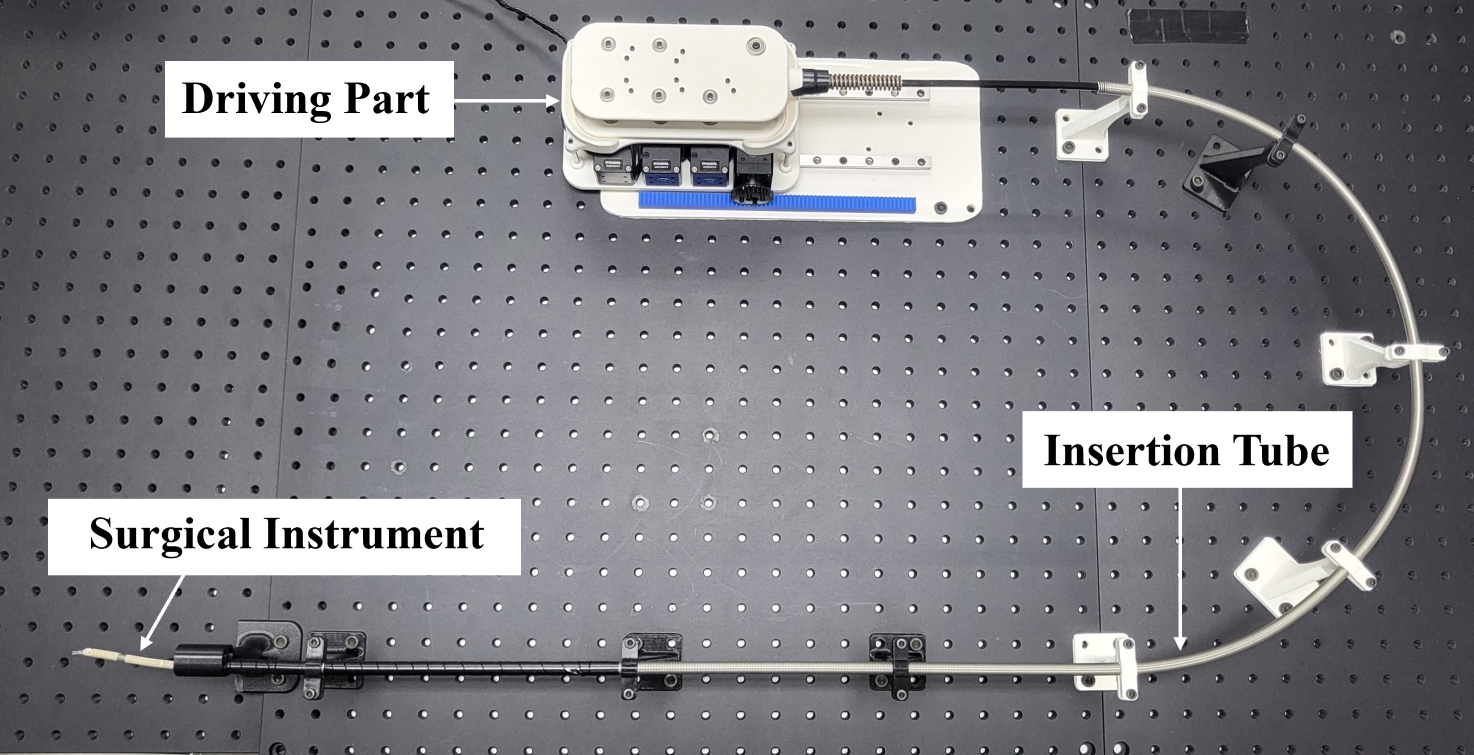}
  \caption{\textbf{Robot Hardware Configuration of the Proposed Continuum Manipulator :} The main components consist of 3 parts: the driving part, insertion tube, and surgical instrument, with a total length of approximately 1.5m.}
  \label{Testbed_Figure}
\end{figure}

The proposed manipulator is consisted with driving parts with 7 actuators, 1.5m length of insertion tube (e.g., connected cables), and extensible continuum instrument as refer to \Cref{Testbed_Figure}. The driving part of the proposed manipulator employs $n$-type actuator, where $n$ represents the number of DOFs \cite{n_type}. In this actuator, two cables are tied to a single motor. {We define the DOFs configuration and cable relationships for the proposed continuum manipulator (refer to \Cref{Cable_Equation}). Definition of each DOFs ($q_{1}$, $q_{2}$, $q_{3}$, $q_{4}$, $q_{5}$, $q_{6}$, and $q_{7}$) are provided (refer to Table I). Specifically, the base portion of the robot’s axial translation ($q_{1}$) and axial rotation ($q_{2}$) is driven by a total of 2 actuators using a rack and pinion and spiral bevel gear mechanism. The extensible  segment has 2 DOFs for pitch and yaw bending ($q_{3}$, $q_{4}$), which are driven by 4 cables. Segment 2 has 1 DOF for pitch bending ($q_{5}$) and is driven by 2 cables. The forceps have 2 DOFs, which include the yaw rotation angle ($q_{6}$) and the grasping angle ($q_{7}$), and are driven by 4 cables (refer to \Cref{Cable_Equation}-(a)). To drive the continuum manipulator with cables, we define the relationship between the continuum segment and the cables. As shown in \Cref{Cable_Equation}-(c), when the segment bends at a specific angle, the cable shape can be approximated as a constant arc. The change in the cable length (\(\Delta c\)) is given by the difference between the central length $h$ and the cable length $c$, as expressed in the following equation:}
{\begin{gather}
\Delta c = h - c
\end{gather}}
{Next, we derive the formulas for the cable length and the radius of curvature $\rho$.
\begin{gather}
c = (\rho - \frac{d}{2})\vartheta, \quad \rho = \frac{h}{\vartheta}
\end{gather}}
{In the equations above and \Cref{Cable_Equation}-(c), $\rho$ is the radius of curvature of the segment, $d/2$ is the offset between the center of the segment and the cable, 
$h$ is the length of the segment, and $c$ is the length of the cable. The final formula for the cable length is as follows:
\begin{gather}
c = \left(\frac{h}{\vartheta} - \frac{d}{2}\right)\vartheta\\
\Delta c = h - \left(\frac{h}{\vartheta} - \frac{d}{2}\right)\vartheta
\end{gather}}
{\noindent The final formula for the change in cable length simplifies as follows, with $h$ being eliminated:
\begin{gather}
\Delta c = \frac{d}{2}\vartheta
\end{gather}}
{Using this equation, the movements for $q_3$, $q_4$, $q_5$, $q_6$, and $q_7$ can be implemented via the cables. However, before driving the cables, the cable coupling issue must be considered. The cable coupling problem arises with the driving cables of segment 2 and the forceps, where bending of the extensible segment or segment 2 can interfere with changes in cable lengths. To resolve this, cable actuation is performed with decoupling, taking into account the bending of the previous segment. Specifically, as shown in \Cref{Cable_Equation}-(d), when the extensible segment bends in the pitch direction, the cables driving segment 2 must adjust to account for the bending of the extensible segment. Similarly, the driving cables for the forceps are also adjusted in response to the bending of both the extensible segment and segment 2. The final relationship can be expressed in matrix form as follows. }
{
\begin{equation}
\therefore \begin{bmatrix}
\Delta m_1 \\
\Delta m_2 \\
\Delta c_3 \\
\Delta c_4 \\
\Delta c_5 \\
\Delta c_6 \\
\Delta c_7 \\
\Delta c_8 \\
\Delta c_9 \\
\Delta c_{10} \\
\Delta c_{11} \\
\Delta c_{12}
\end{bmatrix}
=
\begin{bmatrix}
1 & 0 & 0 & 0 & 0 & 0 & 0 \\
0 & 1 & 0 & 0 & 0 & 0 & 0 \\
0 & 0 & \frac{d_e}{2} & -\frac{d_e}{2} & 0 & 0 & 0 \\
0 & 0 & -\frac{d_e}{2} & \frac{d_e}{2} & 0 & 0 & 0 \\
0 & 0 & \frac{d_e}{2} & \frac{d_e}{2} & 0 & 0 & 0 \\
0 & 0 & -\frac{d_e}{2} & -\frac{d_e}{2} & 0 & 0 & 0 \\
0 & 0 & \frac{d_w}{2} & 0 & \frac{d_{we}}{2} & 0 & 0 \\
0 & 0 & -\frac{d_w}{2} & 0 & -\frac{d_{we}}{2} & 0 & 0 \\
0 & 0 & \frac{d_{jep}}{2} & \frac{d_{jey}}{2} & \frac{d_{jwp}}{2} & \frac{d_j}{2} & -\frac{d_j}{2} \\
0 & 0 & -\frac{d_{jep}}{2} & -\frac{d_{jey}}{2} & -\frac{d_{jwp}}{2} & -\frac{d_j}{2} & \frac{d_j}{2} \\
0 & 0 & -\frac{d_{jep}}{2} & -\frac{d_{jey}}{2} & -\frac{d_{jwp}}{2} & \frac{d_j}{2} & \frac{d_j}{2} \\
0 & 0 & \frac{d_{jep}}{2} & \frac{d_{jey}}{2} & \frac{d_{jwp}}{2} & -\frac{d_j}{2} & -\frac{d_j}{2}
\end{bmatrix}
\begin{bmatrix}
q_1 \\
q_2 \\
q_3 \\
q_4 \\
q_5 \\
q_6 \\
q_7
\end{bmatrix}
\end{equation}}

{Here, $q_1$ and $q_2$ are driven by gears in the actuator, so $m_1$ and $m_2$ represent the motor's rotational angles rather than cable lengths, with a gear ratio of 1:1. The pulley diameter for the forceps is $d_{j}$, and the offsets from the center to all cables are shown in \Cref{Cable_Equation}-(b). Additionally, we assume that the extensible segment maintains a constant bending angle during extension, which implies that, under ideal conditions, there is no change in cable length during extension.}

\section{Hysteresis Analysis and Compensation}
Using an RGBD camera and attached fiducial markers, we obtain the measured physical joint angles ($\textbf{q}_\mathrm{phy}$) corresponding to the commanded joint angles ($\textbf{q}_\mathrm{cmd}$). Through the collected dataset, we trained TCN to model the hysteresis. The trained TCN model estimates the command joint angles ($\textbf{q}_\mathrm{cmd}$) based on the inputted physical joint angles ($\textbf{q}_\mathrm{phy}$). Leveraging the trained TCN models, we propose a hysteresis compensation control strategy to achieve more precise control of the manipulator.

\subsection{{Estimation of Physical Joint Configuration} }
\label{sec:physical}

\begin{figure}[t!]
  \centering
  \includegraphics[width=0.6\linewidth]{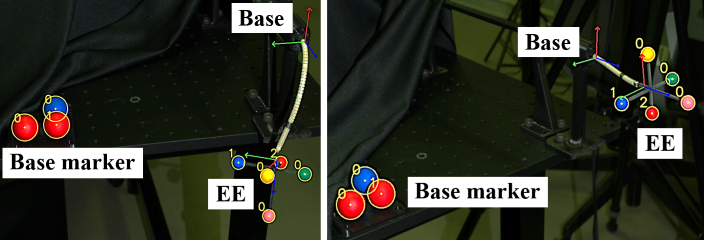}
  \caption{\textbf{Physical Joint Angle Estimation:} Using 8 fiduical markers and RGBD camera, we obtain the transformation matrix of cam to base (${^\mathrm{cam}\mathbf{T}_\mathrm{base}}$), and cam to EE (${^\mathrm{cam}\mathbf{T}_\mathrm{EE}}$).}
  \label{zivid_detection}
\end{figure}

We utilize eight fiducial markers and two RGBD cameras to detect the physical joint angles. As shown in \Cref{zivid_detection}, we employ HSV thresholding to identify each marker and obtain its corresponding point cloud data. The RANSAC algorithm \cite{ransac} is then applied to estimate the center of each marker.
The base pose of the manipulator is calculated using the following equations:
\begin{gather}
^\mathrm{cam}\hat{\textbf{x}}_\mathrm{base} = ^\mathrm{cam}\hat{\textbf{y}}_\mathrm{base} \cross ^\mathrm{cam}\hat{\textbf{z}}_\mathrm{base} \nonumber \\
^\mathrm{cam}\hat{\textbf{y}}_\mathrm{base} = (^\mathrm{cam}\textbf{p}_{r_0} - ^\mathrm{cam}\textbf{p}_{r_1}) / \norm{^\mathrm{cam}\textbf{p}_{r_0} - ^\mathrm{cam}\textbf{p}_{r_1}} \nonumber \\
^\mathrm{cam}\hat{\textbf{z}}_\mathrm{base} = (^\mathrm{cam}\textbf{p}_{r_1} - ^\mathrm{cam}\textbf{p}_{b_0}) / \norm{^\mathrm{cam}\textbf{p}_{r_1} - ^\mathrm{cam}\textbf{p}_{b_0}} \nonumber \\
^\mathrm{cam}\textbf{p}_\mathrm{base} = (^\mathrm{cam}\textbf{p}_{r_0} + ^\mathrm{cam}\textbf{p}_{b_0})/2 + \textbf{p}_\mathrm{offset}
\end{gather}
{where} $^\mathrm{cam}\hat{\textbf{x}}_\mathrm{base}$, $^\mathrm{cam}\hat{\textbf{y}}_\mathrm{base}$, and $^\mathrm{cam}\hat{\textbf{z}}_\mathrm{base}$ are the unit vectors of the robot base frame in the camera frame. $^\mathrm{cam}\textbf{p}_{r_0}$, $^\mathrm{cam}\textbf{p}_{r_1}$, and $^\mathrm{cam}\textbf{p}_{b_0}$ denote the center positions of red ball 0, red ball 1, and blue ball 0 in the camera frame (see \Cref{zivid_detection} for marker index information). $\textbf{p}_\mathrm{offset}$ is the fixed offset between the manipulator base and the base markers, determined from design parameters.

\begin{figure}[t!]
  \centering
  \includegraphics[width=0.65\linewidth]{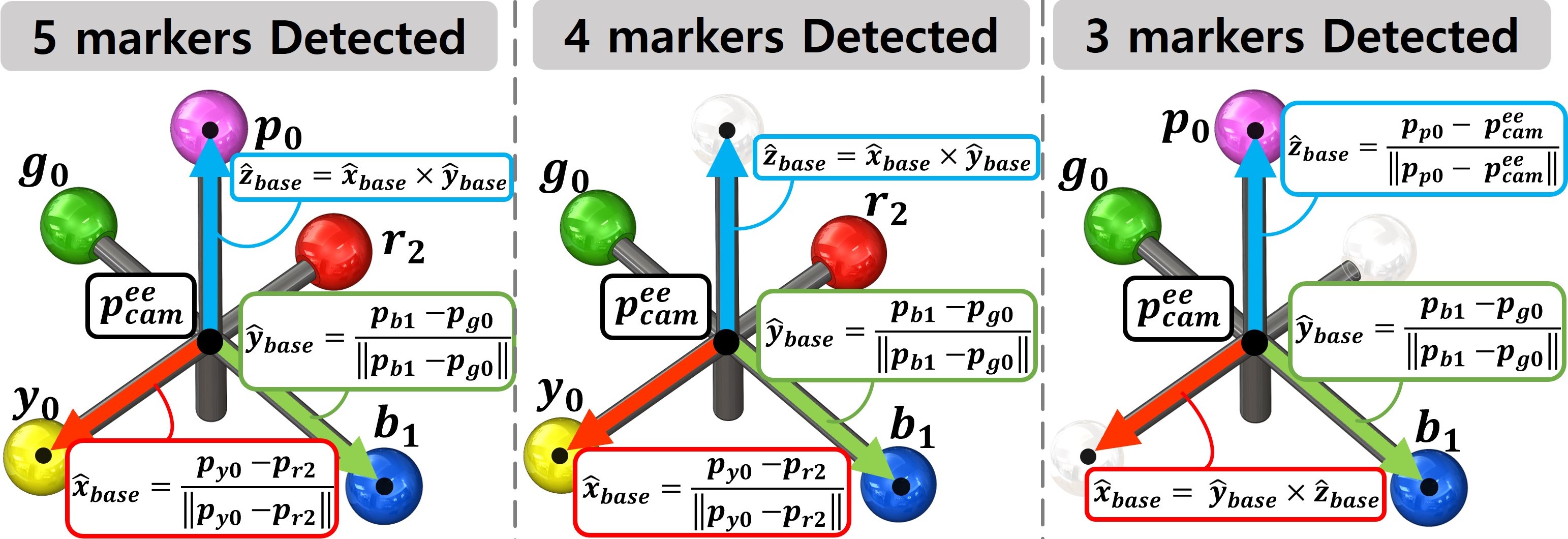}
  \caption{\textbf{Pose Estimation of Manipulator's EE using 5 Fiducial Markers:} This figure illustrates the method used to estimate the relative pose (position and orientation) of the camera frame with respect to the manipulator's EE using five fiducial markers. Three scenarios are considered: (1) All five markers are detected. (2) Four markers are visible due to occlusion. (4 different configurations exist depending on the occluded markers.) (3) Three markers are detected due to occlusion. (total of 10 different configurations exist). For scenarios with partial occlusion (cases 2 and 3), the approach involves obtaining the 3D coordinates (denoted by $\textbf{p}^{EE}_{cam}$) of the EE center in the camera frame using the detected markers. This then allows for calculating the EE frame.}
  \label{end_effector_marker}
\end{figure}

The EE pose detection is described in \Cref{end_effector_marker}. To address potential marker occlusion during data collection, we use five markers instead of the minimum three required to determine the EE pose.

The transformation matrix from the base frame to the EE frame (${^\mathrm{base}\mathbf{T}_\mathrm{ee}}$) is estimated using the obtained camera-to-base (${^\mathrm{cam}\mathbf{T}_\mathrm{base}}$) and camera-to-EE (${^\mathrm{cam}\mathbf{T}_\mathrm{ee}}$) transformation matrices:
\begin{gather}
{^\mathrm{base}\mathbf{T}_\mathrm{ee}}= \: {^\mathrm{cam}\mathbf{T}_\mathrm{base}^{-1}} \cdot {^\mathrm{cam}\mathbf{T}_\mathrm{ee}}   
\end{gather}

With the computed ${^\mathrm{base}\mathbf{T}_\mathrm{ee}}$ matrix, we can solve the inverse kinematics (detailed in \Cref{sec:kinematics}) to obtain the physical joint angles ($\textbf{q}_\mathrm{phy}$).

\subsection{Data Collection and Hysteresis Analysis}
\label{sec:data_collection}

{Building on the physical joint angles obtained in \Cref{sec:physical}, we investigate the hysteresis effects of the continuum manipulator. These effects are induced by various factors, including elongation \cite{extension} (e.g., driving cable), friction \cite{friction} (e.g., between the driving cable and sheath, and between the driving wire and slit joints), twist \cite{twist}, coupling \cite{coupling} (e.g., coupling between proximal and distal segments), cyclic deformation of PEEK \cite{Shrestha2016CyclicDA}, amplified hysteresis due to the 1.5m cable \cite{cable_hystere}, and stiffness changes in the extensible segment due to translation (i.e., as translation increases, the activated length of the extensible segment grows, leading to decreased structural stiffness).}

To analyze this effect, we generated a set of random trajectories (denoted by $\mathcal{D}_\mathrm{trans_i}$). Within each trajectory, the command joint angles for joints $q_2$ to $q_7$ are identical. These angles are randomly chosen within specific ranges for each joint (i.e., $q_2$: $[-30^{\circ}, 30^{\circ}$], $q_3$ to $q_5$ : $[-60^{\circ}, 60^{\circ}]$, $q_6$ to $q_7$ = $0^{\circ}$).

The translation value ($q_1$) varies between trajectories, represented by the subscript $i$ in $\mathcal{D}_\mathrm{trans_i}$. Each command joint angle is linearly interpolated with a step size of 3 degrees. The trajectory data is structured as follows:
\begin{gather}
\mathcal{D}_\mathrm{trans_i} = \{\textbf{q}_\mathrm{cmd}, \textbf{q}_\mathrm{phy}\}_{1}^{N} \nonumber \\
\mathcal{D} = \{\mathcal{D}_\mathrm{trans_i} |(i    = 0, 10, 20, 30, 40, 50 )\}
\end{gather}
\begin{itemize}
    \item $\mathcal{D}_\mathrm{trans_i}$ : This dataset stores pairs of corresponding command joint angles ($\textbf{q}_\mathrm{cmd}$) and detected physical joint angles ($\textbf{q}_\mathrm{phy}$) for $N$ data points.
    \item $\mathcal{D}$ : This dataset combines multiple $\mathcal{D}_\mathrm{trans_i}$ datasets for different translation values ($i$ = 0, 10, 20, 30, 40, 50 mm).
\end{itemize}

\begin{figure}[t!]
  \centering
  \includegraphics[width=0.825\linewidth]{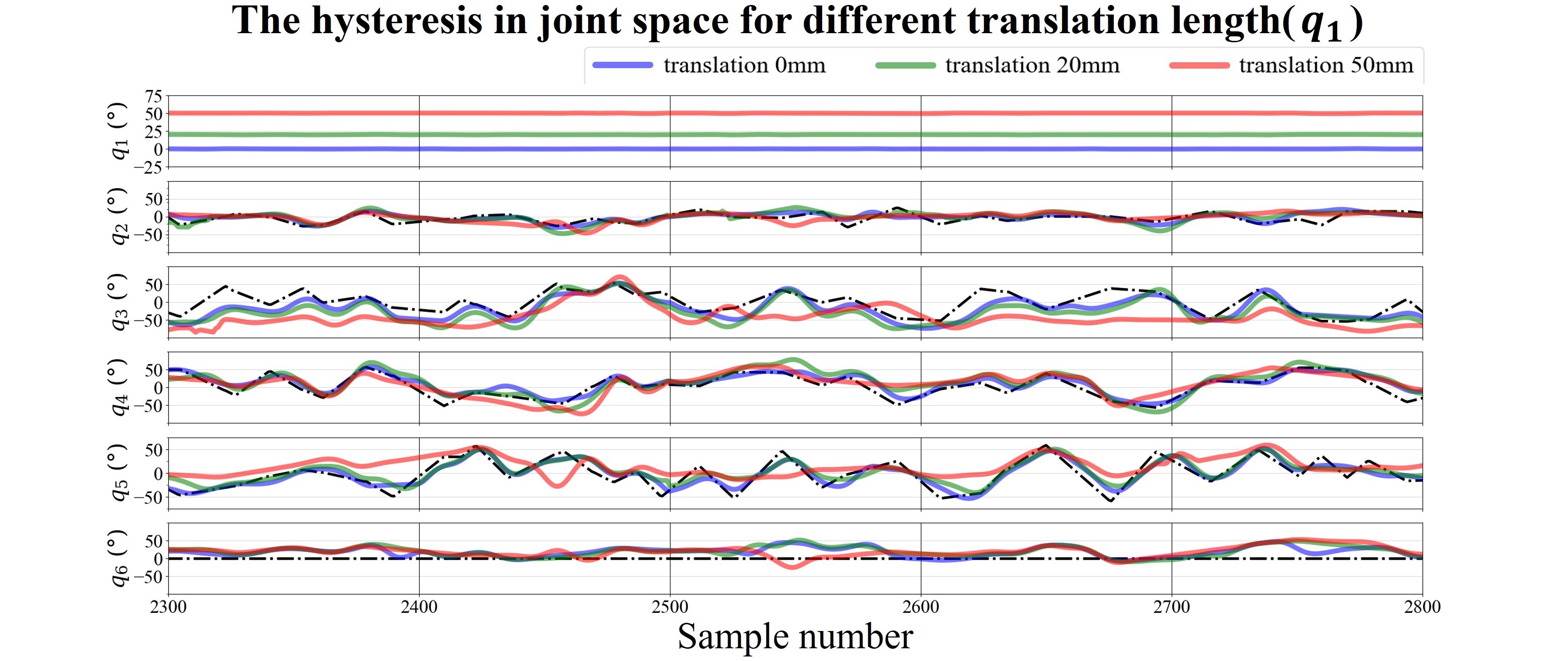}
  \caption{\textbf{The Comparison of the Command and the Physical Joint Angles through the Collected $\mathcal{D}$:} The X-axis represents the sample number during the measurement process. The Y-axis represents the joint angle (in degrees) for each joint of the SAM instrument (denoted as $q_1$, $q_2$, ..., $q_6$). The black line states the $\textbf{q}_\mathrm{cmd}$. The blue, green, and red dashed lines represent the measured physical joint angles ($\textbf{q}_\mathrm{phy}$) on 0mm, 20mm, 50mm translations, respectively. The difference between $\textbf{q}_\mathrm{cmd}$ and $\textbf{q}_\mathrm{phy}$ illustrates the hysteresis effect.}
  \label{hysteresis_line}
\end{figure}

\begin{figure*}[t!]
  \centering
  \includegraphics[width=0.82\linewidth]{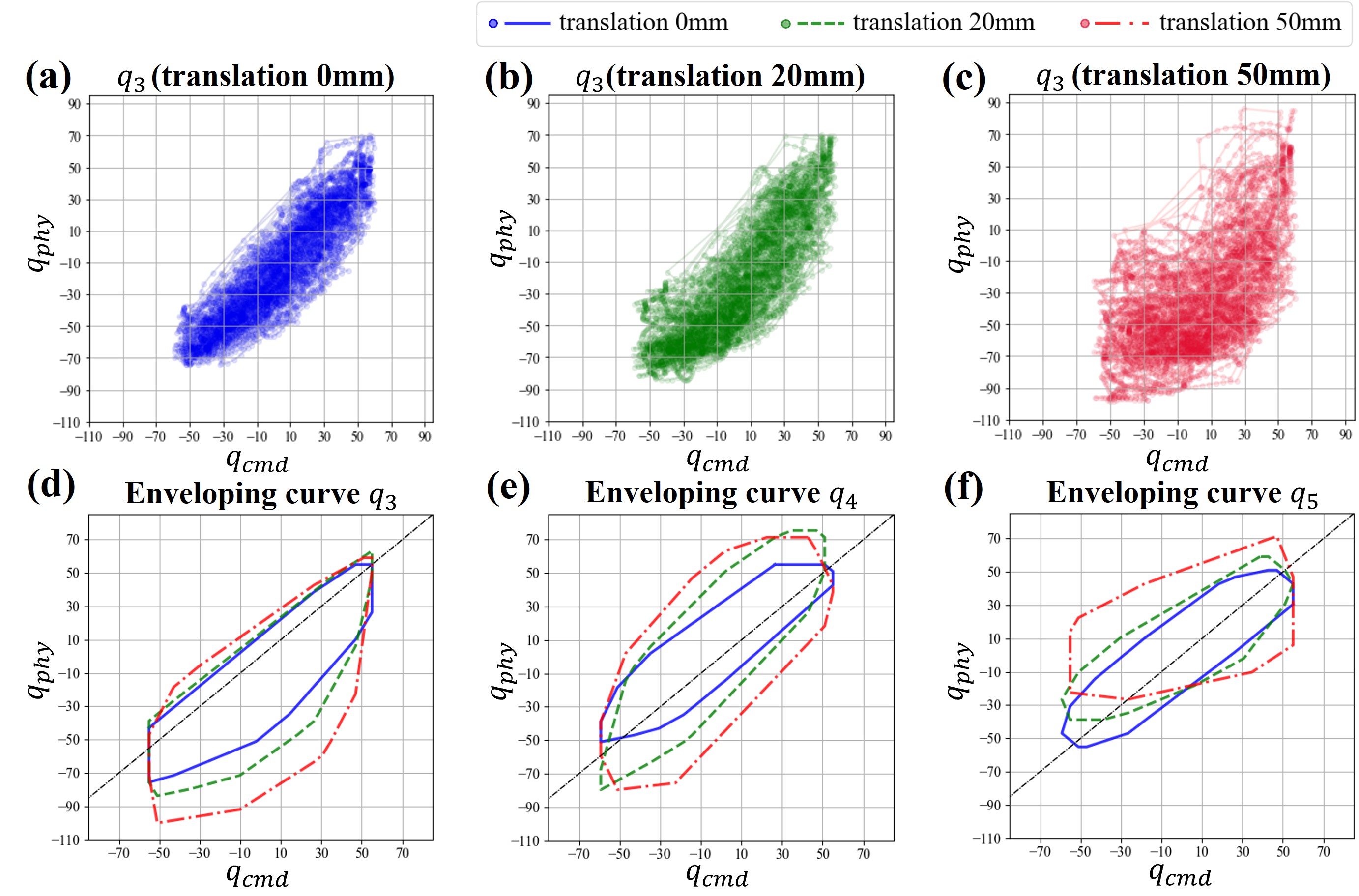}
  \caption{{\textbf{Hysteresis Distribution on Different Translation:}} This figure depicts the hysteresis distribution of the proposed SAM instrument at different extension levels (translation lengths). The blue, green, and red color states translation 0 mm, 20 mm, and 50 mm, respectively. (a), (b), and (c) states the hysteresis distribution on $q_3$ on different translations, and each figure plotted utilizing collected dataset, $\mathcal{D}$. {We observe the history-dependant properties through the connected lines with scatter points.} (d), (e), and (f) state the outermost points of the hysteresis enveloping curve on $q_3$, $q_4$, and $q_5$, respectively. In (d), (e), and (f), the black dashed line represents the reference line. Notice that the $q_3$ exhibit a wider variation as the extension level increases {even though giving same range of $\mathbf{q}_\mathrm{cmd}$.} The proposed SAM mechanisms exhibits different nonlinear hysteresis on different extension levels and higher extension levels lead to a larger discrepancy between commanded and achieved joint angles.}
  \label{hysteresis_curve}
\end{figure*}

We analyzed the hysteresis of the SAM using the collected dataset $\mathcal{D}$ ($N$ = 4,955, total: 29,730). Detailed statistics and figures are presented in \Cref{hysteresis_line} and \Cref{tab:hysteresis_statics}. {The analysis revealed several trends: (1) the physical joint angles, $\mathbf{q}_\mathrm{phy}$, exhibited a rightward shift compared to the command angles, $\mathbf{q}_\mathrm{cmd}$ (see \Cref{hysteresis_line}); (2) hysteresis displayed time-dependent properties—physical joints changed based on previous commands even when the same command was issued (see \Cref{hysteresis_curve}(a), (b), and (c)); (3) coupling effects caused physical joint angles to deviate even when the command for $q_6$ remained constant at 0$^\circ$. Additionally, the Mean Absolute Error (MAE, (23)) of joint angles $q_3$ and $q_4$ increased significantly as the translation distance grew, particularly in the pitch and yaw directions of the extensible segment (see \Cref{hysteresis_curve}(d), (e), and (f)).}

\begin{gather}
    MAE = \frac{1}{n}\sum_{i=1}^{n}\:{\vert \textbf{q}_\mathrm{phy}-\textbf{q}_\mathrm{cmd} \vert} \\
    MSE = \frac{1}{n}\sum_{i=1}^{n}\:(\textbf{q}_\mathrm{phy}-\textbf{q}_\mathrm{cmd})
\end{gather}

Furthermore, $q_3$ exhibited biased hysteresis with a near-equal relationship between the Mean Signed Error (MSE, (23)) and MAE across various translation values (e.g., MAE/MSE of 18.1/15.3 at 0 mm, 28.4/25.7 at 25 mm, and 46.4/42.1 at 50 mm translation). As shown in \Cref{hysteresis_curve}-(a), $\textbf{q}_\mathrm{phy}$ consistently falls below the reference line across all $\textbf{q}_\mathrm{cmd}$ values. This bias stems from the inherent difficulty of $n$-type actuators in maintaining equal initial tension in antagonistic cable pairs. The initial tension imbalance is amplified by larger translations, which reduce structural stiffness, resulting in higher deflections at greater extension levels. We infer that the cable driving positive $q_3$ likely has a higher initial tension than its negative counterpart, contributing to this observed bias. 
{In summary, our observations indicate that several factors contribute to hysteresis in the SAM, most notably the structural stiffness reduction induced by translation. Additionally, we observed wide dead zones, likely caused by friction between joints and the driving cable, friction between the cable and sheath, and the material properties of PEEK (see Appendix \Cref{hysteresis repeat}). Furthermore, coupling effects were evident, as physical joint values deviated even when command values remained constant. Lastly, initial tension differences introduced during the manufacturing process may further exacerbate hysteresis. Given these complexities, we believe that modeling this nonlinear hysteresis through analytical methods presents significant challenges.}

\begin{table}[b!]
\centering
\renewcommand{\arraystretch}{1.2} 
\caption{Hysteresis statics on different translation}
\label{tab:hysteresis_statics}
\begin{tabular}{cc|cccccc} 
 \toprule
 \multirow{4}{*}{} Translation & Error & \begin{tabular}[c]{@{}c@{}} $q_1$\\ (mm) \end{tabular} & \begin{tabular}[c]{@{}c@{}} $q_2$\\ ($^\circ$) \end{tabular} & \begin{tabular}[c]{@{}c@{}} $q_3$\\ ($^\circ$) \end{tabular} & \begin{tabular}[c]{@{}c@{}} $q_4$\\ ($^\circ$) \end{tabular} & \begin{tabular}[c]{@{}c@{}} $q_5$\\ ($^\circ$) \end{tabular} & \begin{tabular}[c]{@{}c@{}} $q_6$\\ ($^\circ$) \end{tabular} \\  
 \midrule \midrule
 \multirow{4}{*}{0mm} & MAE & 0.1 & 8.2 & \textbf{18.1} & \textbf{11.9} & 10.3 & 19.0 \\ 
 & SD & 0.1 & 6.0 & 12.0 & 9.1 & 7.4 & 12.8\\
 & MSE & 0.0 & 1.0 & \textbf{15.3} & -7.7 & -0.1 & -17.4\\
 & SD & 0.1 & 10.1 & 15.4 & 12.9 & 12.7 & 14.8\\
 \hline
 \multirow{4}{*}{20mm} & MAE & 0.1 & 11.2 & \textbf{28.4} & \textbf{18.3} & 12.6 & 19.6 \\ 
 & SD & 0.1 & 7.9 & 16.5 & 12.4 & 9.4 & 13.5\\
 & MSE & 0.0 & 0.1 & \textbf{25.7} & -9.2 & -5.6 & -17.9\\
 & SD & 0.2 & 13.7 & 20.4 & 20.1 & 14.7 & 15.7\\
 \hline
 \multirow{4}{*}{50mm} & MAE & 0.1 & 15.1 & \textbf{46.4} & \textbf{30.0} & 21.7 & 23.1\\
 & SD & 0.1 & 10.7 & 24.9 & 19.8 & 15.3 & 16.0\\
 & MSE & 0.0 & -0.1 & \textbf{42.1} & -10.8 & -16.4 & -17.8\\ 
 & SD & 0.1 & 18.5 & 31.6 & 34.3 & 20.9 & 21.7\\
 \bottomrule
\end{tabular}
\end{table}

\subsection{Hysteresis Modeling using Deep Learning}
\label{deep_learning}

{As demonstrated in Appendix \Cref{hysteresis repeat}, the SAM exhibits repeatable hysteresis. This repeatability is observed in single-joint hysteresis, and even in random trajectories where all joints move simultaneously. Despite the nonlinear and complex nature of SAM’s hysteresis, this repeatability enables the modeling of hysteresis using deep learning methods.} 

\begin{figure*}[t!]
  \centering
  \includegraphics[width=0.75\linewidth]{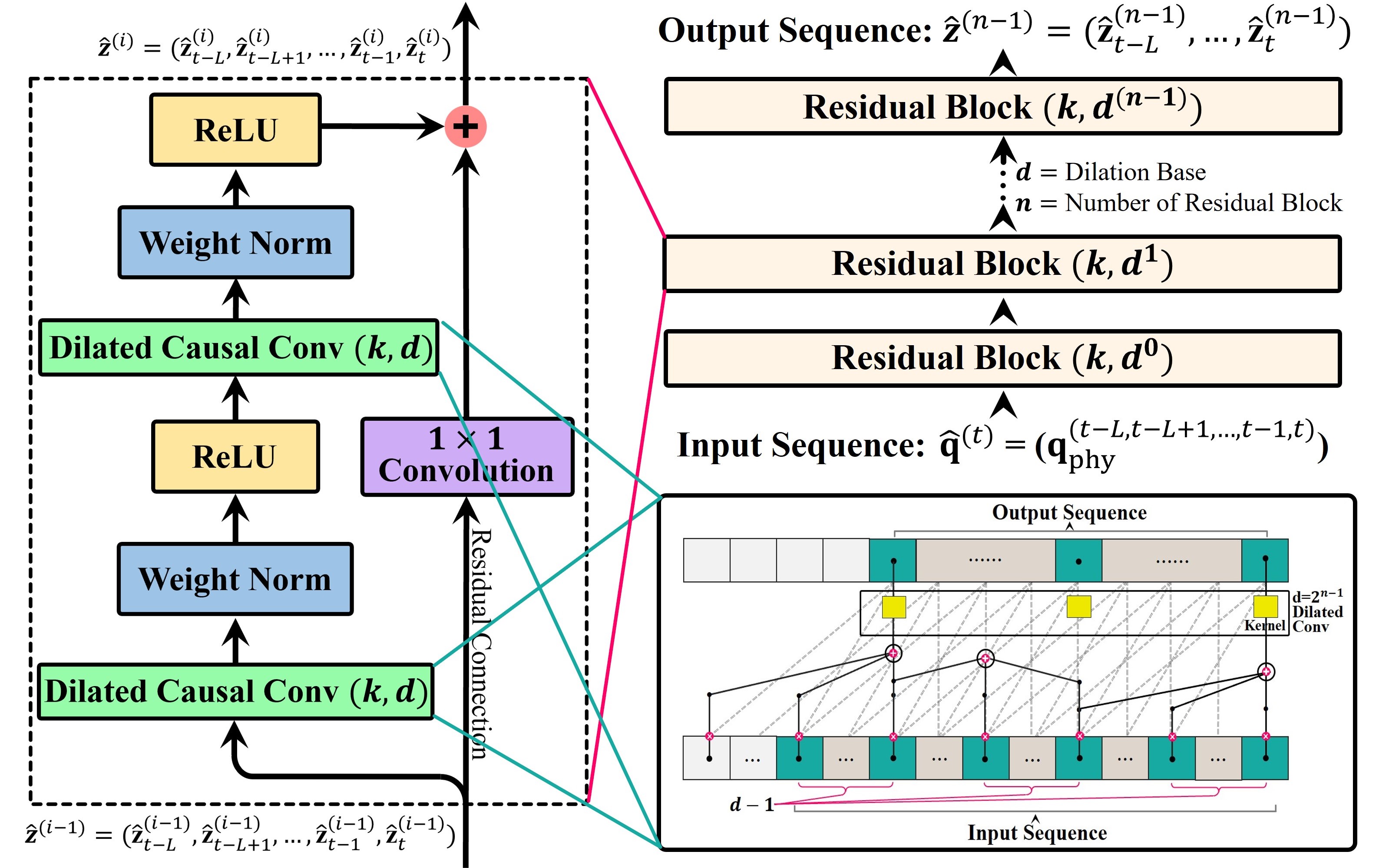}
  \caption{\textbf{Architecture of TCN:} TCN composed with layers of 1D dilated convolution layer, which can get features from the history of the joints. The dilation base was set as $2$, so the dilation factor is increase in $2^{0}$, $2^{1}$, ..., and $2^{i}$. The last of the feature vector, $\hat{z}_{L}^t$, become the outputs of the model.}
  \label{tcn_architecture}
\end{figure*}

We employed a TCN \cite{tcn} for hysteresis estimation due to its proven effectiveness in this domain. As detailed in \cite{park2024hysteresis}, TCN efficiently compensate for the hysteresis effect. The TCN architecture consists of serially connected residual blocks. Each residual block incorporates two dilated convolutions, two weight normalization layers, and two ReLU activation functions (refer to Fig. \ref{tcn_architecture} for details). The residual blocks employ exponentially increasing dilation factors with a base of 2 ($d$). The first residual block has dilation factor of $1$($2^0$), the second residual block has dilation factor of $2$ ($2^1$), and others has $2^{n-1}$ dilation factor. The number of residual blocks ($n$) are determined by the (24). In (24), the $L$ is the input sequence length and $k$ is kernel size. In our setting, the kernel size is 3. The TCN returns the feature vectors of the input sequence ($\hat{\textbf{z}}^{(n-1)}$). The last one of the feature vector ($\hat{\textbf{z}}_t^{(n-1)}$) serving as the estimated $\textbf{q}_\mathrm{cmd}$ corresponding to the input $\textbf{q}_\mathrm{phy}$ as refer to (25).
\begin{gather}
num \: block = \lceil \log_{2} \frac{(L-1)}{2k-2} + 1 \rceil \\
\hat{\textbf{q}}_\mathrm{cmd}^{(t)} = \hat{\textbf{z}}_{t}^{(n-1)} = f_{\theta} (\textbf{q}_\mathrm{phy}^{(t-L, \: t-L+1, \:..., \:t-1, \:t)})
\end{gather}

For training the TCN models, we utilize a dataset, $\mathcal{D}_\mathrm{train}$, containing 4,955 randomly commanded joint angles across 6 different translations, resulting in a total of 29,730 data points. Validation datasets, $\mathcal{D}_\mathrm{valid}$, are employed, each containing 1,307 command joint angles across 6 translations, totaling 7,842 data points.

\begin{table}[b!]
\centering
\caption{Performance comparison of TCN-inverse models for \\each sequence length, $L$ on test dataset}
\label{tab:model_performance}
\begin{tabular}{ccccc} 
\toprule
MAE / L                                   & $L = 10$ & $L = 50$ & $L = 100$ & $L = 150$  \\ 
\midrule
\begin{tabular}[c]{@{}c@{}}Model 1\\$[MAE \pm SD]$\end{tabular} & $5.5\pm8.3$ &   $7.6\pm10.3$   & $8.3\pm11.0$ &    $8.5\pm11.1$    \\ 
\midrule
\begin{tabular}[c]{@{}c@{}}Model 2\\$[MAE \pm SD]$\end{tabular} & $5.6\pm8.3$ &  $8.1\pm11.9$  &  $8.8\pm12.0$   &    $9.7\pm12.7$    \\ 
\midrule
\begin{tabular}[c]{@{}c@{}}Model 3\\$[MAE \pm SD]$\end{tabular} & $5.6\pm8.4$ &  $8.6\pm11.8$  &   $9.0\pm12.0$  &   $9.8\pm13.3$     \\
\toprule
\end{tabular}
\end{table}

To mitigate biases from random weight initialization, each TCN model is trained three times with varying initial weights. Additionally, we investigate the impact of different input sequence lengths ($L$ = 10, 50, 100, 150) and determine the optimal input sequence length which can capture the history-dependent hysteresis in continuum manipulators.

During training, we use a fixed learning rate of 0.001, mean squared error as the loss function, and utilization of Adam optimizer. After 10,000 epochs, the model with the lowest validation loss from each training phase is selected. These optimal models are then evaluated on unseen trajectories from the test dataset ($\mathcal{D}_{test}$).

As detailed in Table~\ref{tab:model_performance} (e.g., Mean Absolute Error (MAE) and Standard Deviation (SD)), the TCN model with a sequence length ($L$) of 10 achieve the best performance, exhibiting lower MAE and SD compared to other lengths. This suggests that a memory of the past 10 timesteps is sufficient for the model to effectively capture the hysteresis behavior.

\subsection{Design of Hysteresis Compensation Algorithm}
\label{sec:compensation_algo}
In this section, we present the design of the proposed hysteresis compensation algorithm, which leverages three TCN with input sequence length $L = 10$. These models achieved the best performance on the test dataset, as detailed in Section ~\ref{deep_learning}. Hysteresis compensation aims to return calibrated command joint angles that can accurately reach the inputted desired joint angles. 
The trained TCN models (refer to (25)) predict command joint angles for given physical joint angles. Consequently, the predicted command joint angles corresponding to the desired joint angles can be directly input into the control algorithm. Our proposed hysteresis compensation control algorithm exclusively employs TCN models. These models receive a sequence of desired joint angles ($\textbf{q}_\mathrm{desired}$) as input and output the corresponding calibrated commanded joint angles ($\textbf{q}_\mathrm{cal}$).

\begin{figure}[t!]
  \centering
  \includegraphics[width=0.55\linewidth]{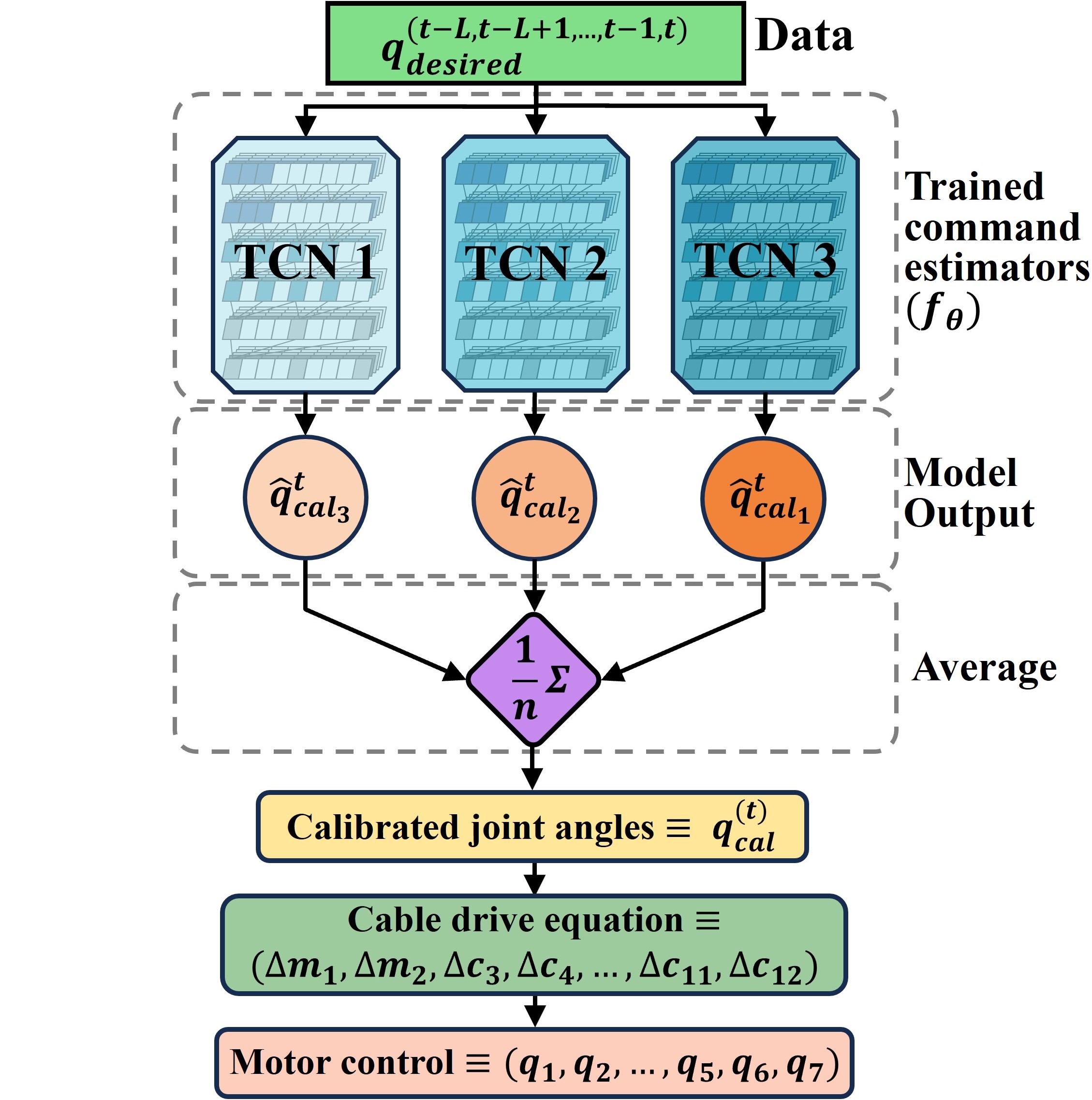}
  \caption{\textbf{Hysteresis Compensation Control Method:} Using the trained TCN models, we get the output, $\hat{\textbf{q}}_\mathrm{{cal_{1,2,3}}}^{(t)}$, on the desired joint angle. Then, we get average of the $\hat{\textbf{q}}_\mathrm{{cal_{1,2,3}}}^{(t)}$, to obtain the final calibrated joint angle ($\hat{\textbf{q}}_\mathrm{cal}$).}
  \label{control_loop}
  \vspace{-1em}
\end{figure}

As shown in \Cref{algo1} and \Cref{control_loop}, the calibrated command joint angle ($\hat{\textbf{q}}_\mathrm{cal}^{(t)}$) is computed as the average of outputs from three individual TCN models ($\hat{\textbf{q}}_\mathrm{cal_1}^{(t)}$, $\hat{\textbf{q}}_\mathrm{cal_2}^{(t)}$, $\hat{\textbf{q}}_\mathrm{cal_3}^{(t)}$). This ensemble approach aims to enhance output stability. Due to random weight initialization, the three models exhibit some variance even with identical inputs. Averaging their outputs mitigates this effect and improves accuracy. The proposed algorithm achieves a time latency of 1ms, ensuring real-time control capabilities.

\begin{algorithm}[b!]
\caption{Hysteresis Compensation Algorithm}
\label{algo1}
\begin{algorithmic}[1]
    \REQUIRE Sequence of desired joint angles $\textbf{q}_\mathrm{desired}^{t, ..., t-M}$ $(M \leq L)$, command estimators $f_{\theta_1}$, $f_{\theta_2}$, and $f_{\theta_3}$, time sequence length $L$
    \IF {$M$ $<$ $L$}
        \STATE $\textbf{q}_\mathrm{desired}^{t-M-1},..., \textbf{q}_\mathrm{desired}^{t-L}$ $\leftarrow zero$ $padding$  
    \ENDIF
    \STATE $\hat{\textbf{q}}_\mathrm{cal_{1,2,3}}^{(t)} = f_{\theta_{1,2,3}}(\textbf{q}_\mathrm{desired}^{(t)},..., \textbf{q}_\mathrm{desired}^{(t-L)})$
    \STATE $\textbf{q}_\mathrm{cal}^{(t)} = (\hat{\textbf{q}}_\mathrm{cal_1}^{(t)} + \hat{\textbf{q}}_\mathrm{cal_2}^{(t)} + \hat{\textbf{q}}_\mathrm{cal_3}^{(t)}) / 3$
\end{algorithmic}
\end{algorithm}
\section{Results and Validation}
\label{results_valid}
This section evaluates the performance of the calibrated controller in comparison to the uncalibrated controller.
\begin{itemize}
\item Calibrated Controller: The desired joint angles ($\textbf{q}_\mathrm{desired}$) serve as input. The calibrated command joint angles ($\textbf{q}_\mathrm{cal}$) for achieving $\textbf{q}_\mathrm{desired}$ are computed by processing them through the three trained TCN models (as described in \Cref{sec:compensation_algo}). Subsequently, the motor commands for $\textbf{q}_\mathrm{desired}$ are calculated using (25).
\item Uncalibrated Controller: The desired joint angles are directly input into the control equation (18).
\end{itemize}
The validation process comprises two distinct tasks: a random trajectory tracking test and a box pointing task. These tasks are designed to assess the accuracy of the calibrated control in both joint space (via the random trajectory tracking test in \Cref{random_trajectory}) and operational space (via the box pointing task in \Cref{box_pointing_test}).

\subsection{Random Trajectory Tracking Test}
\label{random_trajectory}
This section compares the performance of the uncalibrated and calibrated controllers in tracking random joint space trajectories. It's important to note that the training data in Section~\ref{deep_learning} included translations at $0$, $10$, $20$, $30$, $40$, and $50 mm$. To test generalization, random trajectories are generated for unseen translations: $5 mm$, $25 mm$, and $45 mm$.
\begin{figure*}[t!]
  \centering
  \includegraphics[width=\linewidth]{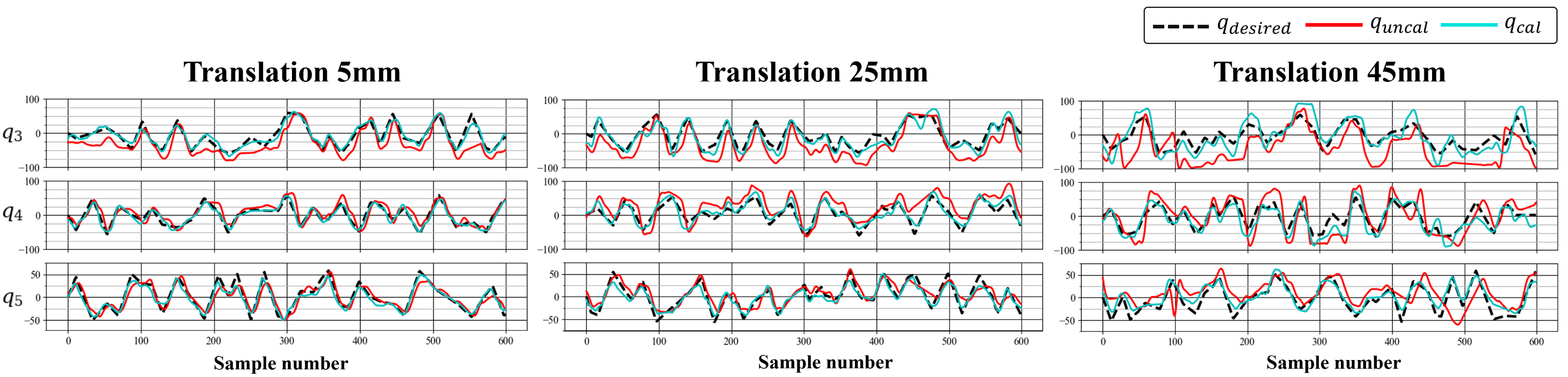}
  \caption{\textbf{Performance Comparison of Calibrated and Uncalibrated Controller on Random Trajectory Tracking Test using Three Unseen Trajectory:} This figure compares the performance of a calibrated controller (blue line, $\textbf{q}_\mathrm{cal}$ ) and an uncalibrated controller (red line, $\textbf{q}_\mathrm{uncal}$ ) on a random trajectory tracking test using three unseen trajectories (not previously encountered by the controllers). The black dashed line represents the desired trajectory ($\textbf{q}_\mathrm{desired}$ ). The calibrated controller shows significantly better tracking performance, following the desired trajectory more closely compared to the uncalibrated controller. This highlights the effectiveness of the calibration process in improving the accuracy of the manipulator's movements. }
  \label{fig:cal_uncal}
\end{figure*}

\begin{table*}[t!]
\centering
\caption{Performance Comparison between Uncalibrated and Calibrated Control on Random Trajectory Tracking Test}
\label{cal_vs_uncal}
\label{tab:statics}
\begin{tabular}{llccccc|ccccc|ccccc} 
\toprule
\multicolumn{2}{c}{\multirow{2}{*}{\begin{tabular}[c]{@{}l@{}}Translation\\~~~~/ MAE \\\end{tabular}}} & \multicolumn{5}{c}{\textbf{Translation 5 mm}}                             & \multicolumn{5}{c}{\textbf{Translation 25 mm}}                              & \multicolumn{5}{c}{\textbf{Translation 45 mm}}                                \\
\multicolumn{2}{l}{}                                                                          & $q_2$        & $q_3$        & $q_4$        & $q_5$        & $q_6$         & $q_2$        & $q_3$         & $q_4$         & $q_5$        & $q_6$         & $q_2$         & $q_3$         & $q_4$         & $q_5$        & $q_6$          \\ 
\midrule
\multirow{2}{*}{\textbf{Caibrated}}   & MAE                                                         & \textbf{4.2} & \textbf{6.8} & \textbf{6.6} & \textbf{6.3} & \textbf{14.2} & \textbf{8.5} & \textbf{14.6} & \textbf{11.0} & \textbf{7.5} & \textbf{17.0} & \textbf{12.2} & \textbf{18.3} & \textbf{17.7} & \textbf{9.5} & \textbf{16.3}  \\
                                & SD                                                          & 3.1          & 5.4          & 5.2          & 5.1          & 9.2           & 6.1          & 11.9          & 8.3           & 5.7          & 11.7          & 8.2           & 13.8          & 12.6          & 7.7          & 12.3           \\ 
\midrule
\multirow{2}{*}{\textbf{Uncalibrated}} & MAE                                                         & 8.9          & 22.3         & 12.4         & 9.4          & 19.7          & 13.3         & 31.2          & 25.3          & 10.7         & 24.6          & 15.5          & 43.3          & 29.6          & 20.7         & 26.7           \\
                                & SD                                                          & 6.0          & 14.7         & 9.3          & 6.7          & 11.4          & 8.9          & 16.4          & 14.6          & 8.3          & 14.9          & 10.6          & 23.8          & 19.6          & 15.2         & 18.1           \\ 
\bottomrule
\multicolumn{17}{l}{* MAE represents that mean absolute error.}                                                                                                                                                                                                                          \\
\end{tabular}
\end{table*}
For joints $q_3$ to $q_5$, values are randomly assigned within the range $[-60^{\circ} : 60^{\circ}]$, while for $q_2$, values are randomly assigned within $[-30^{\circ} :30^{\circ}]$. Each trajectory comprises 955, 887, and 941 desired joint angles ($\textbf{q}_\mathrm{desired}$) respectively, with each angle linearly interpolated by 3°.

We obtained the physical joint angles for both the uncalibrated and calibrated controllers using an RGBD camera and 8 fiducial markers (employing the same methods as in Section~\ref{sec:physical}), with $\textbf{q}_\mathrm{desired}$ as input. The results for physical joint angles of each controller are detailed in \Cref{fig:cal_uncal} for the three trajectories (translations: $5 mm$, $25 mm$, and $45 mm$).
The proposed calibrated controller effectively mitigates hysteresis effects, consistently demonstrating lower Mean Absolute Error (MAE) across all joints compared to the uncalibrated controller (refer to \Cref{fig:cal_uncal} and \Cref{cal_vs_uncal}). Notably, these significant improvements are achieved even with unseen translations, highlighting the generalizability of the compensation method beyond the specific training data. The summary of the notable observed reductions in MAE is as follows:

\begin{itemize}
    \item Translation 5 mm: $q_3$, $q_4$, $q_5$ exhibit reductions of 69.5 \%, 46.8 \%, and 33.0 \%, respectively.
    \item Translation 25 mm: $q_3$, $q_4$, $q_5$ exhibit reductions of 53.2\%, 56.5\%, and 29.9\%, respectively.
    \item Translation 45 mm: $q_3$, $q_4$, $q_5$ exhibit reductions of 57.7\%, 40.2\%, and 54.1\%, respectively.
\end{itemize} 

\subsection{Box Pointing Task}
\label{box_poinitng_section}
\begin{figure}[t!]
  \centering
  \includegraphics[width=0.45\linewidth]{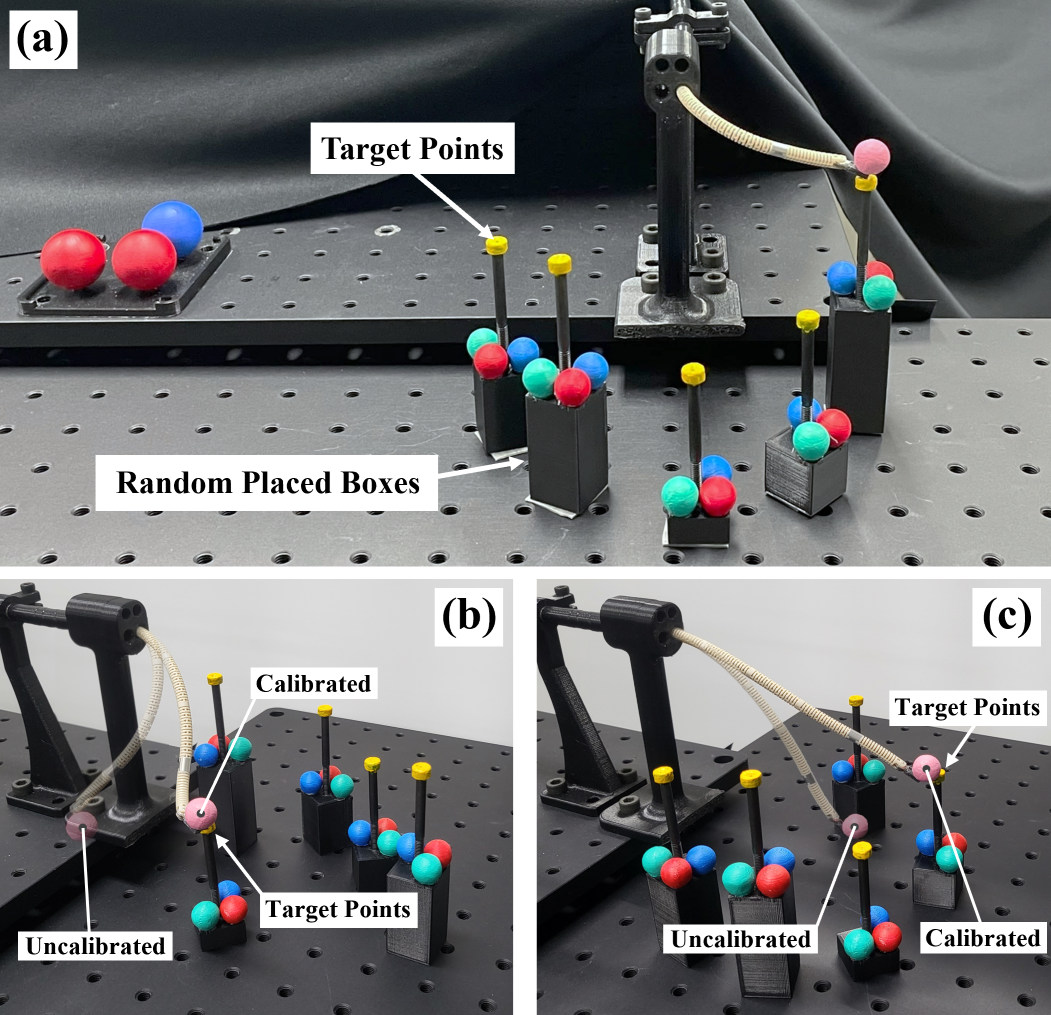}
  \caption{\textbf{{Box Pointing Task :}} (a) Demonstration figure of the task: This part of the evaluation assesses the effectiveness of the EE in precisely reaching randomly positioned target points (yellow) on five boxes. We compare the positioning error between the calibrated and uncalibrated control approaches. (b) and (c) visualize the difference in positioning error between the calibrated and uncalibrated controllers relative to the target points.}
  \label{box_pointing_test}
  \vspace{-1em}
\end{figure}

This section assesses the effectiveness of the proposed hysteresis compensation method in achieving accurate positioning. The experiment utilizes five boxes of varying heights, each attached with unique red, green, and blue fiducial markers {(refer to \Cref{box_pointing_test}).} {Additionally, a pink single ball marker is used to verify the position of manipulator's EE. This marker is much lighter, weighing less than 1g, compared to the 4g marker (refer to \Cref{end_effector_marker}) used in the data collection in \Cref{sec:data_collection}. This is intended to validate the compensation for hysteresis while accounting for the deflection caused by the marker’s weight applied in the data collection}.

An RGBD camera captures the markers, enabling the estimation of their centers using the methods described in Section~\ref{sec:physical}. These marker positions are then used to compute the transformation matrix from the camera frame to each box frame ${^{cam}\mathbf{T}_{box_j}}$ $(j=1,2,3,4,5)$ using (26):
\begin{gather}
^\mathrm{cam}\hat{\textbf{x}}_\mathrm{box_j} = ^\mathrm{cam}\hat{\textbf{y}}_\mathrm{box_j}\cross ^\mathrm{cam}\hat{\textbf{z}}_\mathrm{box_j} \nonumber \\
^\mathrm{cam}\hat{\textbf{y}}_\mathrm{box_j} = (^\mathrm{cam}\textbf{p}_\mathrm{r_j} - ^\mathrm{cam}\textbf{p}_{g_j}) / \norm{^\mathrm{cam}\textbf{p}_\mathrm{r_j} - ^\mathrm{cam}\textbf{p}_{g_j}} \nonumber \\
^\mathrm{cam}\hat{\textbf{z}}_\mathrm{box_j} = (^\mathrm{cam}\textbf{p}_{r_j} - ^\mathrm{cam}\textbf{p}_{b_j}) / \norm{^\mathrm{cam}\textbf{p}_\mathrm{r_j} - ^\mathrm{cam}\textbf{p}_{b_j}} \nonumber \\
^\mathrm{cam}\textbf{p}_\mathrm{box_j} = (^\mathrm{cam}\textbf{p}_{g_j} + ^\mathrm{cam}\textbf{p}_{b_j})/2 + \textbf{p}_\mathrm{x_{offset}}
\end{gather}
where $^\mathrm{cam}\textbf{p}_{r_j}$, $^\mathrm{cam}\textbf{p}_{b_j}$, and $^\mathrm{cam}\textbf{p}_{g_j}$ are the center positions (in the camera frame) of the red, blue, and green spheres on each box (denoted by subscript j), respectively. Using inverse kinematics (detailed in Section~\ref{sec:kinematics}), the target joint angles ($\textbf{q}_\mathrm{box_j}$) required to reach designated points on each box are calculated through the obtained ${^\mathrm{cam}\mathbf{T}_\mathrm{box_j}}$.
The manipulator executes a series of motions to reach target positions on each box, beginning with the lowest box and sequentially progressing to the highest. This process is repeated 15 times with the boxes placed in random configurations. Two control strategies are compared: the uncalibrated controller and the proposed calibrated controller. The position errors between the desired positions and the physical EE's positions are evaluated for both controllers.

\begin{table}[b!]
\centering
\caption{Performance Comparison between Calibrated and \\Uncalibrated Control on Box Picking Task}
\label{tab:box}
\begin{tabular}{cccc} 
\toprule
\multicolumn{2}{c}{\begin{tabular}[c]{@{}c@{}}EE \\ Position Error\end{tabular}}             & \begin{tabular}[c]{@{}c@{}}Calibrated\\Control\end{tabular} & \begin{tabular}[c]{@{}c@{}}Uncalibrated\\Control\end{tabular}  \\ 
\toprule
\multirow{2}{*}{$x$ (mm)}                                                           & MAE & \textbf{13.31}                                              & 21.09                                                          \\
                                                                                  & SD  & \textbf{15.19}                                              & 24.57                                                          \\ 
\midrule
\multirow{2}{*}{$y$ (mm)}                                                           & MAE & \textbf{12.22}                                              & 19.73                                                          \\
                                                                                  & SD  & \textbf{14.55}                                              & 27.86                                                          \\ 
\midrule
\multirow{2}{*}{$z$ (mm)}                                                           & MAE & \textbf{11.50}                                              & 20.04                                                          \\
                                                                                  & SD  & \textbf{14.47}                                              & 23.12                                                          \\ 
\midrule
\multirow{2}{*}{\begin{tabular}[c]{@{}c@{}}Euclidean\\distance (mm)\end{tabular}} & AVG & \textbf{23.99}                                              & 32.33                                                          \\
                                                                                  & SD  & \textbf{9.52}                                               & 21.95                                                          \\
\bottomrule
\end{tabular}
\end{table}

The proposed calibrated controller demonstrates consistent superiority over the uncalibrated controller across all spatial dimensions ($x$, $y$, $z$) and the overall Euclidean distance. As detailed in Table~\ref{tab:box}, the calibrated controller achieves significant error reductions:

\begin{itemize}
\item $x:(mm)$: MAE and SD are decreased by 36.9\% and 38.2\%, respectively.
\item $y:(mm)$: MAE and SD are decreased by 38.1\% and 47.8\%, respectively.
\item $z:(mm)$: MAE and SD are decreased by 43.5\% and 37.4\%, respectively.
\item Euclidean Distance: MAE and SD are decreased by 25.8\% and 56.7\%, respectively.
\end{itemize}

\section{CONCLUSIONS}

This study introduced a novel continuum manipulator incorporating SAM, enabling extended reach without increasing size or DOFs. The proposed design demonstrated a significant 527.6\% extension in reachable workspace volume compared to conventional continuum manipulators. However, the instrument exhibited considerable hysteresis, primarily due to cable effects such as friction, elongation, and coupling. Moreover, the hysteresis model was observed to vary with increasing extension length, attributed to changes in structural stiffness. 

To address the variation in hysteresis with extension, we proposed a real-time deep learning-based compensation control algorithm. We utilized an RGBD camera and 8 fiducial markers to collect joint angle data from the manipulator. Through collected dataset, we trained TCN models. These models estimate the command joint angles for the inputted physical joint angles. Trajectory tracking tests on unseen trajectories demonstrated significant and consistent reduction in hysteresis across all joint angles, particularly in the highly affected $q_3$ joint. The calibrated controller achieved substantial improvements in joint space: for translation 5mm, $q_3$ error was reduced from 22.3° to 6.8°; for translation 25mm, $q_3$ error went from 31.2° to 14.6°; and for translation 45mm, $q_3$ error decreased from 43.3° to 18.3°. Similarly, the box pointing task with the calibrated controller showed significant reductions in position error across all axes: $x$ (mm) error decreased from 21.09 to 13.31 (MAE), $y$ (mm) error went from 19.73 to 12.22 (MAE), $z$ (mm) error reduced from 20.04 to 11.32 (MAE), and the average Euclidean distance error decreased from 32.33mm to 23.99mm. These results demonstrate that, despite the presence of hysteresis, the proposed TCN-based compensation controller effectively reduced both error and deviation. 

Our research implies that in future surgical applications, the proposed SAM mechanism could enable surgical tools to access various lesions without requiring access to the overtube, potentially minimizing damage to surrounding tissues. Additionally, the improvement in hysteresis compensation has the potential to significantly enhance surgical task performance by reducing position and joint angle errors in real-time.

{However, as outlined in Appendix \Cref{sec:deflection_marker}, the use of fiducial markers may cause deflection in the continuum manipulator, introducing additional complexity. Specifically, modeling hysteresis using marker-based data collection might inadvertently capture the gravitational effects on the manipulator. Since current markerless detection methods exhibit larger detection errors compared to the RGBD camera system (e.g., RGBD camera’s position error is 0.34 ± 0.18 mm \cite{Hwang_Dvrk}), we opted to use the RGBD system in our study to more accurately capture hysteresis. Nonetheless, our compensation methods remain applicable across different detection techniques, as the hysteresis is modeled using command and detected physical joint angles. Therefore, even with markerless pose estimation, we expect our methods to effectively compensate for hysteresis. In future work, we aim to further improve accuracy by minimizing detection errors and addressing the issue of deflection.}

\section{ACKNOWLEDGEMENT}
This work was supported by the DGIST R\&D Program of the Ministry of Science and ICT (23-PCOE-02, 23-DPIC-20), by the DGIST Start-up Fund Program of the Ministry of Science and ICT (2024010213), and by the collaborative project with ROEN Surgical Inc. This work was supported by the Korea Medical Device Development Fund grant funded by the Korea government (Project Number: 1711196477 , RS-2023-00252244) and by the National Research Council of Science \& Technology (NST) grant funded by the Korea government (MSIT) (CRC23021-000).

\bibliographystyle{IEEEtran}
\bibliography{ref}

\newpage

\section{Appendix}
\label{appendix:appendix}
\addcontentsline{toc}{section}{Appendices}
\renewcommand{\thesubsection}{\Alph{subsection}}

\subsection{Terminology, Coordinate Frame and Modeling Assumptions} 
\label{appendix:term}

\begin{table}[b!]
\centering
\caption{Nomenclature for Extensible Continuum Manipulator}
\label{parameter_table}
\begin{tabular}{ll} 
\toprule
\rule{0pt}{5ex}
\raisebox{1.5ex}{\textbf{Symbol}} & \raisebox{1.5ex}{\textbf{Definition}} \\ \toprule
Frame Index ($i$) & \begin{tabular}[c]{@{}l@{}}Sequential numbering of frames from the base to the EE of the manipulator.\end{tabular} \\ 
\midrule
Bending Angle ($\varphi$) & \begin{tabular}[c]{@{}l@{}}Angle between the bending plane and a reference axis.\end{tabular} \\ 
\midrule
Curvature ($k_x$) & Curvature around the x-axis. \\ 
\midrule
Curvature ($k_y$) & Curvature around the y-axis. \\ 
\midrule
Curvature ($k$) & \begin{tabular}[c]{@{}l@{}}Overall curvature along the bent segment from its base to the end.\end{tabular} \\ 
\midrule
Arc Length ($s_{1}$) & \begin{tabular}[c]{@{}l@{}}Arc length of the bent extensible segment from its base to the end.\end{tabular} \\ 
\midrule
Arc Length ($s_{2}$) & \begin{tabular}[c]{@{}l@{}}Arc length of segment 2 from its base to the end.\end{tabular} \\ 
\midrule
Central Length ($l_1$) & The central length of the semi-active segment. \\ 
\midrule
\begin{tabular}[c]{@{}l@{}}Translation Length ($q_1$)\end{tabular} & \begin{tabular}[c]{@{}l@{}}Translation distance (mm) along the z-axis from the base.\end{tabular} \\ 
\midrule
\begin{tabular}[c]{@{}l@{}}Rotation Angle ($q_2$)\end{tabular} & \begin{tabular}[c]{@{}l@{}}Angle of rotation in the roll direction at the base.\end{tabular} \\ 
\midrule
\begin{tabular}[c]{@{}l@{}}Bending Angle of \\Extensible Segment ($q_3, q_4$)\end{tabular} & \begin{tabular}[c]{@{}l@{}}Angles of bending in the pitch and yaw directions of extensible segment.\end{tabular} \\ 
\midrule
\begin{tabular}[c]{@{}l@{}}Bending Angle of \\Segment2 ($q_5$)\end{tabular} & \begin{tabular}[c]{@{}l@{}}The angle by bending in the pitch direction of segment 2.\end{tabular} \\ 
\midrule
\begin{tabular}[c]{@{}l@{}}Rotation Angle of \\Forceps ($q_6$)\end{tabular} & \begin{tabular}[c]{@{}l@{}}The angle by rotating in the yaw direction from forceps.\end{tabular} \\ 
\midrule
\begin{tabular}[c]{@{}l@{}}Grasping Angle of \\Forceps ($q_7$)\end{tabular} & \begin{tabular}[c]{@{}l@{}}The angle by rotating in the yaw direction from forceps.\end{tabular} \\ 
\midrule
\begin{tabular}[c]{@{}l@{}}Angle of \\Forceps1 ($q_{f_1}$)\end{tabular} & \begin{tabular}[c]{@{}l@{}}The angle of forceps2.\end{tabular} \\ 
\midrule
\begin{tabular}[c]{@{}l@{}}Angle of Forceps2 ($q_{f_2}$)\end{tabular} & \begin{tabular}[c]{@{}l@{}}The angle of forceps1.\end{tabular} \\ 
\midrule
Bending angle ($\vartheta$) & The total bending angle of each segment. \\ 
\midrule
\begin{tabular}[c]{@{}l@{}}Homogeneous \\Transformation Matrix ($^{i-1}{\mathbf{T}}_{i}$)\end{tabular} & \begin{tabular}[c]{@{}l@{}}The homogeneous transformation matrix from $i-1$ to $i$ of extensible segment.\end{tabular} \\ 
\midrule
\begin{tabular}[c]{@{}l@{}}Rotation Matrix ($\mathbf{R}_{z}\left(\varphi \right )$)\end{tabular} & \begin{tabular}[c]{@{}l@{}}Rotation matrix that rotates by $\varphi$ about the z-axis.\end{tabular} \\ 
\midrule
\begin{tabular}[c]{@{}l@{}}Rotation Matrix ($\mathbf{R}_{x}\left(\kappa s \right )$)\end{tabular} & \begin{tabular}[c]{@{}l@{}}Rotation matrix that rotates by $\kappa s$ about the x-axis.\end{tabular} \\ 
\midrule
\begin{tabular}[c]{@{}l@{}}Rotation Matrix ($\mathbf{R}_{z}\left ( -\varphi \right )$)\end{tabular} & \begin{tabular}[c]{@{}l@{}}Rotation matrix that rotates by $\varphi$ about the z-axis.\end{tabular} \\ 
\midrule
\begin{tabular}[c]{@{}l@{}}Translation Vector ($^{i-1}\mathbf{\textrm{P}}_{i}$)\end{tabular} & \begin{tabular}[c]{@{}l@{}}Translation vector from $i-1$ to $i$.\end{tabular} \\ 
\bottomrule
\end{tabular}
\end{table}

For kinematics modeling, we employ the piecewise constant-curvature approximation, treating each segment of the continuum manipulator as an arc with uniform curvature. In the proposed continuum segment model, both the extensible segment and segment 2 utilize these coordinate frames to determine the position and orientation of their EEs. The terminology and coordinate frames used are illustrated in \Cref{kinematics_Figure}, and a detailed description of the coordinate frames is as follows:\\

\begin{figure}[t!]
  \centering
  \includegraphics[width=0.9\linewidth]{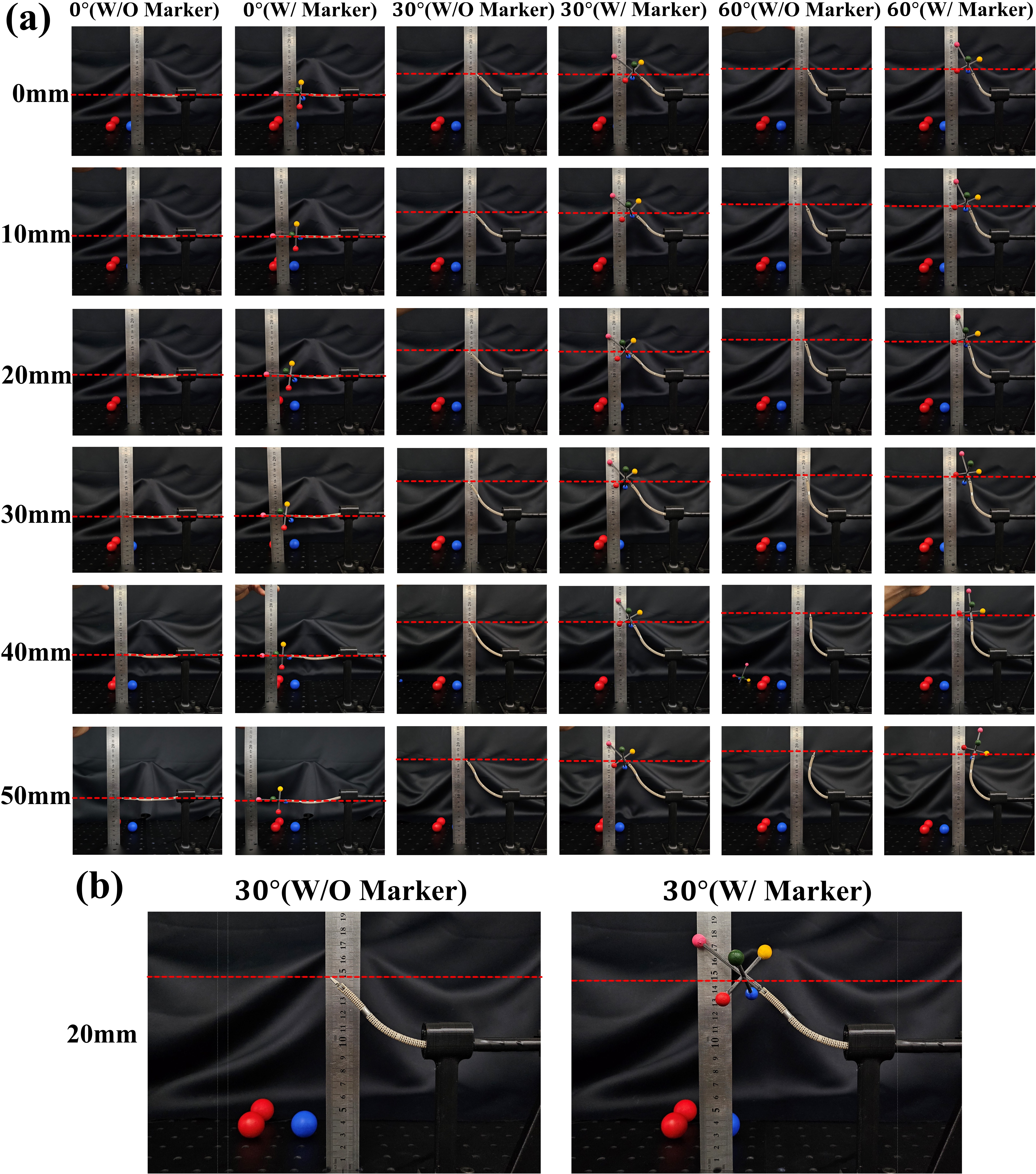}
  \caption{\textbf{{Photographs of SAM Continuum Manipulator EE with and without Markers at Varying Translation Lengths and Bending Angles :}} {Images are captured at translation lengths of 0mm, 10mm, 20mm, 30mm, 40mm, and 50mm for bending angles of 0°, 30°, and 60°, both with (W/) and without (W/O) markers attached. A ruler is placed alongside to provide a clear measurement reference. The red line indicates the position of the SAM continuum manipulator’s EE.}}
  \label{Deflection_picture}
\end{figure}

\begin{itemize}
    \item Base coordinate frame of the continuum segment ($ O_{i-1} \equiv \left \{ {\hat{x}_{i-1}},{\hat{y}_{i-1}},{\hat{z}_{i-1}} \right \}$): The origin of $O_{i-1}$ is positioned at the center of the segment's base. The axis $z_{i-1}$ is oriented perpendicular to the base plane.\\
    
    \item Base Coordinate frame of the continuum segment in bending plane (${O}'_{i-1} \equiv \left \{ {{\hat{x}}'_{i-1}},{{\hat{y}}'_{i-1}},{{\hat{z}}'_{i-1}} \right \}$): The origin of ${O}'_{i-1}$ aligns with that of $O_{i-1}$. ${O}'_{i-1}$ is derived from $O_{i-1}$ through a rotation by $\varphi$ about the z-axis of $O_{i-1}$, aligning the frame with the segment's bending plane.\\

    \item End coordinate frame of the continuum segment in bending plane (${O}'_{i} \equiv \left \{ {{\hat{x}}'_{i}},{{\hat{y}}'_{i}},{{\hat{z}}'_{i}} \right \}$): The origin of ${O}'_{i}$ is located at the segment's end center. Its ZY plane is parallel to the bending plane of the segment.\\

    \item End coordinate frame of the continuum segment ($ O_{i} \equiv \left \{ {\hat{x}_{i}},{\hat{y}_{i}},{\hat{z}_{i}} \right \}$): The origin of $O_{i}$ aligns with that of ${O}'_{i}$. $O_{i}$ is derived from ${O}'_{i}$ through a rotation by $-\varphi$ about the z-axis of $O_{i}$. \\
\end{itemize}

The overall local coordinate frame definitions for the proposed continuum manipulator are depicted in \Cref{kinematics_Figure}-(b) and detailed in \Cref{parameter_table}. These frames are established based on the DOFs of the continuum manipulator (refer to \Cref{Cable_Equation}) and are used for setting up the coordinate system to determine the transformation matrix of the final EE.\\

\begin{itemize}
    \item Frame of manipulator base ($O_{1_{b}} \equiv \left \{ {\hat{x}_{1_{b}}},{\hat{y}_{1_{b}}},{\hat{z}_{1_{b}}} \right \}$): The origin of $O_{1_{b}}$ is positioned at the center of the manipulator's base. The axis $z_{1_{b}}$ is oriented perpendicular to the base plane.\\

    \item Frame of Semi-active segment base ($O^{'}_{1_{b}} \equiv \left \{ {{\hat{x}}'_{1_{b}}},{{\hat{y}}'_{1_{b}}},{{\hat{z}}'_{1_{b}}} \right \}$): The origin of $O'_{1_{b}}$ coincides with that of $O_{1_{b}}$. $O'_{1_{b}}$ is rotated by $q_2$ about the z-axis of $O_{b}$.\\

    \item Frame of Semi-active segment end ($O_{1_{e}} \equiv \left\{ \hat{x}_{1_{e}}, \hat{y}_{1_{e}}, \hat{z}_{1_{e}} \right\}$): The origin of $O_{1_{e}}$ is located at the center of the segment's end, and the axis $z_{i-1}$ is oriented perpendicular to the distal plane of the Semi-active segment. $O_{1_{e}}$ results from the Semi-active segment bending by an angle of $\sqrt{q_3^2 + q_4^2}$ and changing the arc length by $q_1$.\\

    \item Frame of segment 2 base ($O_{2_{b}} \equiv \left \{ {\hat{x}_{2_{b}}},{\hat{y}_{2_{b}}},{\hat{z}_{2_{b}}} \right \}$): The origin of $O_{2_{b}}$ is translated along the z-axis by the length of the connector from $O_{1_{e}}$, maintaining the same orientation. \\

    \item Frame of segment 2 end ($O_{2_{e}} \equiv \left \{ {\hat{x}_{2_{e}}},{\hat{y}_{2_{e}}},{\hat{z}_{2_{e}}} \right \}$): The origin of $O_{2_{e}}$ is positioned at the center of the end of segment 2. The axis $z_{2_{e}}$ is oriented perpendicular to the distal plane of segment 2. $O_{2_{e}}$ results from segment 2 bending by an angle of $q_5$. \\

    \item Frame of the manipulator's EE ($O_{ee} \equiv \left \{ {\hat{x}_{ee}},{\hat{y}_{ee}},{\hat{z}_{ee}} \right \}$): The orientation of $O_{ee}$ is obtained by rotating around the $x_{2_{e}}$ axis of $O_{2_{e}}$ by the forceps' yaw rotation angle ($q_6$). The position of $O_{ee}$ is translated along the $z_{2_{e}}$ axis of $O_{2_{e}}$ by the length of the forceps. \\
\end{itemize}

\subsection{Deflection of Manipulator by using Marker}
\label{sec:deflection_marker}
The continuum manipulator’s accuracy can be affected by external factors such as the weight of attached markers, which may cause deflection and impact the detection of joint angles. To assess this impact, we conducted a series of experiments to quantify the deflection resulting from the use of markers during the manipulation tasks.
\begin{table}
\centering
\caption{{Measured Position Z of SAM Continuum Manipulator EE with and without Markers \\at Different Translation Lengths and Bending Angles}}
\label{Deflection_Table}
\resizebox{0.65\linewidth}{!}{%
\renewcommand{\arraystretch}{1.3} 
\begin{tabular}{cccc} 
\toprule
\multicolumn{1}{c}{\multirow{2}{*}{$q_1$ (mm) / $q_3$ ($^\circ$)}} & \multicolumn{3}{c}{position Z [cm] of Without $\rightarrow$ With marker (deflection)}  \\ 
\cmidrule{2-4}
\multicolumn{1}{l}{}        & 0         & 30        & 60         \\ 
\midrule
0    & 10.0 $\rightarrow$ 10.0 (0) & 14.0 $\rightarrow$ 13.9 (0.1) & 14.9 $\rightarrow$ 14.8 (0.2)     \\
10   & 9.7 $\rightarrow$ 9.6 (0.1)   & 14.5 $\rightarrow$ 14.3 (0.2) & 15.9 $\rightarrow$ 15.8 (0.1)   \\
20   & 10.0 $\rightarrow$ 9.7 (0.3) & 14.8 $\rightarrow$ 14.6 (0.2) & 16.7 $\rightarrow$ 16.7 (0)      \\
30   & 9.7 $\rightarrow$ 9.4 (0.3)  & 16.3 $\rightarrow$ 16.1 (0.2) & 17.7 $\rightarrow$ 17.5 (0.2)    \\
40   & 10.0 $\rightarrow$ 9.6 (0.4) & 16.6 $\rightarrow$ 16.3 (0.3) & 18.2 $\rightarrow$ 17.9 (0.3)    \\
50   & 9.6 $\rightarrow$ 9.0 (0.6)  & 17.2 $\rightarrow$ 16.9 (0.3) & 19.0 $\rightarrow$ 18.7 (0.3)    \\
\bottomrule
\end{tabular}
}
\end{table}
We attached markers (approximately 4g) on the continuum manipulator to measure the deflection at different translation lengths (0mm, 10mm, 20mm, 30mm, 40mm, and 50mm) for bending angles of 0°, 30°, and 60°. The experiments were performed under two conditions: with and without markers attached. \Cref{Deflection_picture} and \Cref{Deflection_Table} illustrate the observed deflection patterns under these conditions. Despite the relatively low mass of the markers, some deflection was observed across all tested scenarios. Notably, the maximum deflection was recorded at a translation length of 50mm with a 0° bending angle, reaching up to 6mm. This suggests that even minimal marker weight can cause measurable deflection in the proposed manipulator, particularly when the translation length is extended. 

We recognize that future advancements in markerless pose estimation algorithm hold great potential for further reducing the impact of deflection. Current state-of-the-art markerless pose estimation algorithm are rapidly evolving, yet they continue to face challenges related to estimation errors. For example, the tracking average position and orientation errors are reported as 1.24 ± 0.85 mm and 3.25 ± 1.45°, respectively \cite{changzhou2024, wang2023}. In contrast, the errors induced by manipulator hysteresis are significantly higher, averaging 7.97 ± 1.45 mm and 22.36 ± 4.47°. These estimation inaccuracies can adversely affect the precision of hysteresis compensation algorithms, particularly when modeling hysteresis using deep learning.

In this study, we observed that the accuracy of Ball detection using RGBD cameras (e.g., position error of ball detection \cite{Hwang_Dvrk} = 0.32 ± 0.18mm) is relatively higher compared to markerless pose estimation methods (e.g., position error of markerless detection \cite{changzhou2024, wang2023} = 1.24 ± 0.85 mm). This suggests that Ball detection could capture the hysteresis behavior despite the manipulator’s deflection caused by marker weight. To validate this, in \Cref{box_poinitng_section}, we conducted a box pointing task using a single ball marker weighing less than 1g to assess the ability to compensate for hysteresis despite the marker's weight. The results showed that using the Single ball marker in the box pointing task reduced the Euclidean Distance's MAE and SD by 25.8\% and 56.7\%, respectively, compared to uncalibrated control. This indicates that hysteresis can be effectively captured even when considering the deflection caused by the marker.

Furthermore, our future research will focus on integrating improved markerless approaches to address these minor deflection issues. This integration is expected to enhance the accuracy of hysteresis compensation in continuum manipulators, significantly improving the system's robustness. This integrated approach builds upon the foundation established by our current work and represents a critical step toward improving the practical applicability of continuum manipulators. 

This section highlights the considerations and mitigation strategies regarding deflection caused by markers, indicating that it is possible to maintain the accuracy of the continuum manipulator even with the use of physical markers. Future research will focus on exploring markerless pose detection methods to address the remaining issues.

\subsection{Hysteresis Loop and Repeatability}
\label{hysteresis repeat}

\begin{figure}[t!]
  \centering
  \includegraphics[width=0.98\linewidth]{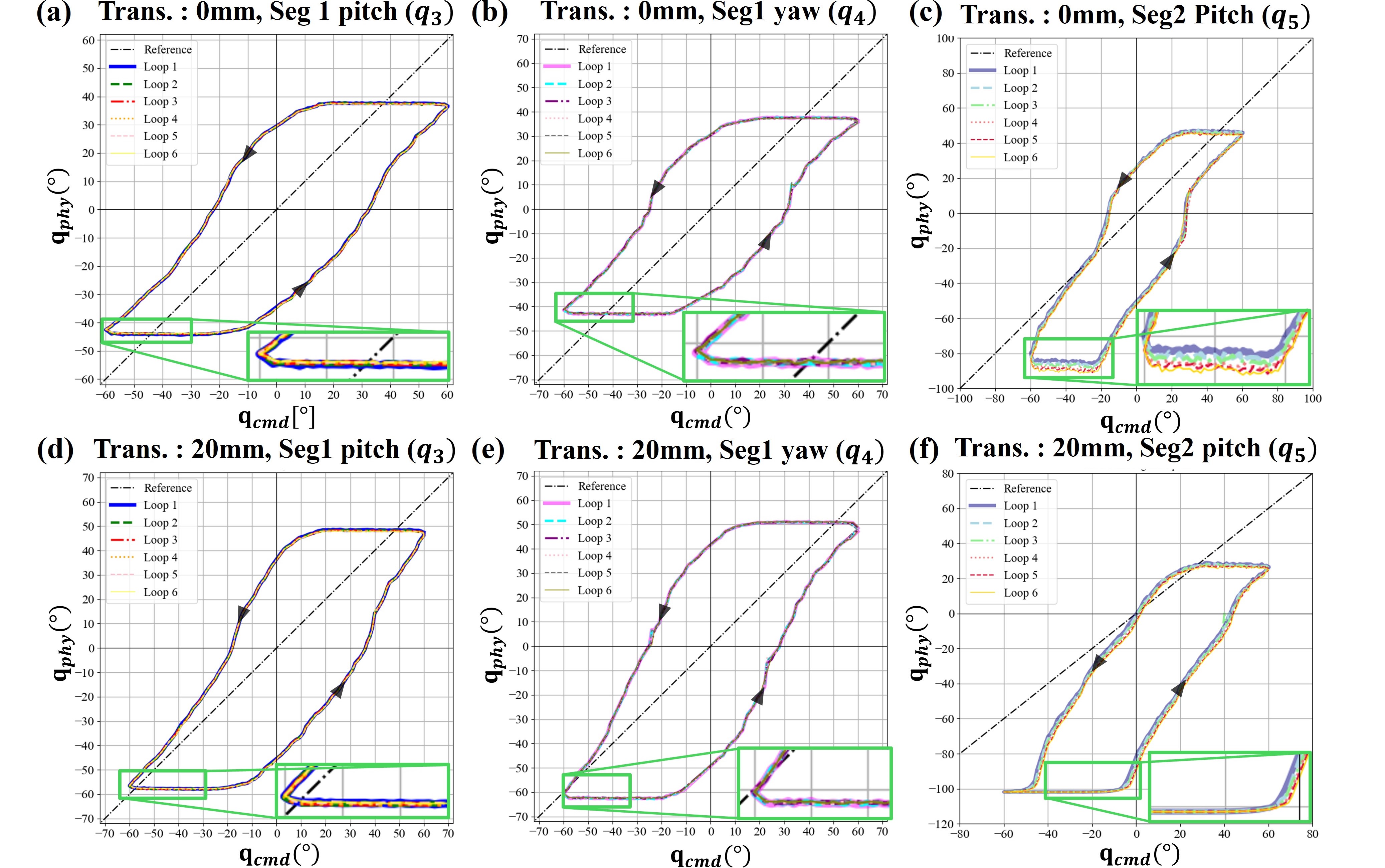}
  \caption{\textbf{Hysteresis Loops of Single DOFs ($\mathbf{q_3}$, $\mathbf{q_4}$, $\mathbf{q_5}$) at Different Translation Distances (0 mm, 20 mm) and Repeatability Validation across Loops}: For each single joint ($q_3$, $q_4$, $q_5$), we commanded repeated joint movements between -60$^\circ$ and 60$^\circ$ while keeping all other joints fixed, forming the corresponding hysteresis loops. (a) - (c) display the hysteresis loops at 0 mm translation for $q_3$, $q_4$, and $q_5$, respectively, and (d) - (f) show the loops at 20 mm translation. The experiment was repeated six times for each joint, and as shown in the enlarged portions of the figure, the loops exhibit high repeatability with minimal deviation across trials. Additionally, noticeable dead zones were observed in all loops.}
  \label{hysteresis_loop_validation}
\end{figure}

\begin{table}[t!]
\centering
\caption{Average and standard deviation of MAE values between loops \\to assess repeatability}
\label{loop_repeatability}
\resizebox{0.55\linewidth}{!}{%
\begin{tabular}{llll}
\toprule
                 & \multicolumn{1}{c}{$q_3$ ($^\circ$)} & \multicolumn{1}{c}{$q_4$ ($^\circ$)} & \multicolumn{1}{c}{$q_5$ ($^\circ$)}  \\ 
\midrule
Translation 0 mm  & 0.16 $\pm$ 0.04           & 0.23 $\pm$ 0.02           & 1.39 $\pm$ 0.60            \\
Translation 20 mm & 0.20 $\pm$ 0.05           & 0.23 $\pm$ 0.038          & 1.29 $\pm$ 0.65            \\
\toprule
\end{tabular}
}
\end{table}

In this section, we examine the hysteresis loops for the single degrees of freedom (DOFs) corresponding to the $q_3, q_4$, and $q_5$ joints under different translation length. For plotting the hysteresis loops in \Cref{hysteresis_loop_validation}, all joints were fixed except for the joint being analyzed (e.g., $q_3$, $q_4$, and $q_5$). The command sequence ($0^\circ  \rightarrow 60^\circ \rightarrow 0^\circ \rightarrow -60^\circ \rightarrow 0^\circ$) was executed seven times for each joint. Using fiducial markers and RGBD sensing, the EE transformation $\mathbf{^{base}T_{ee}}$ was obtained, and the corresponding physical joint angles, $\mathbf{q_{phy}}$ were determined through inverse kinematics (refer to equations (11) and (12)). 

\begin{figure}[t!]
  \centering
  \includegraphics[width=0.925\linewidth]{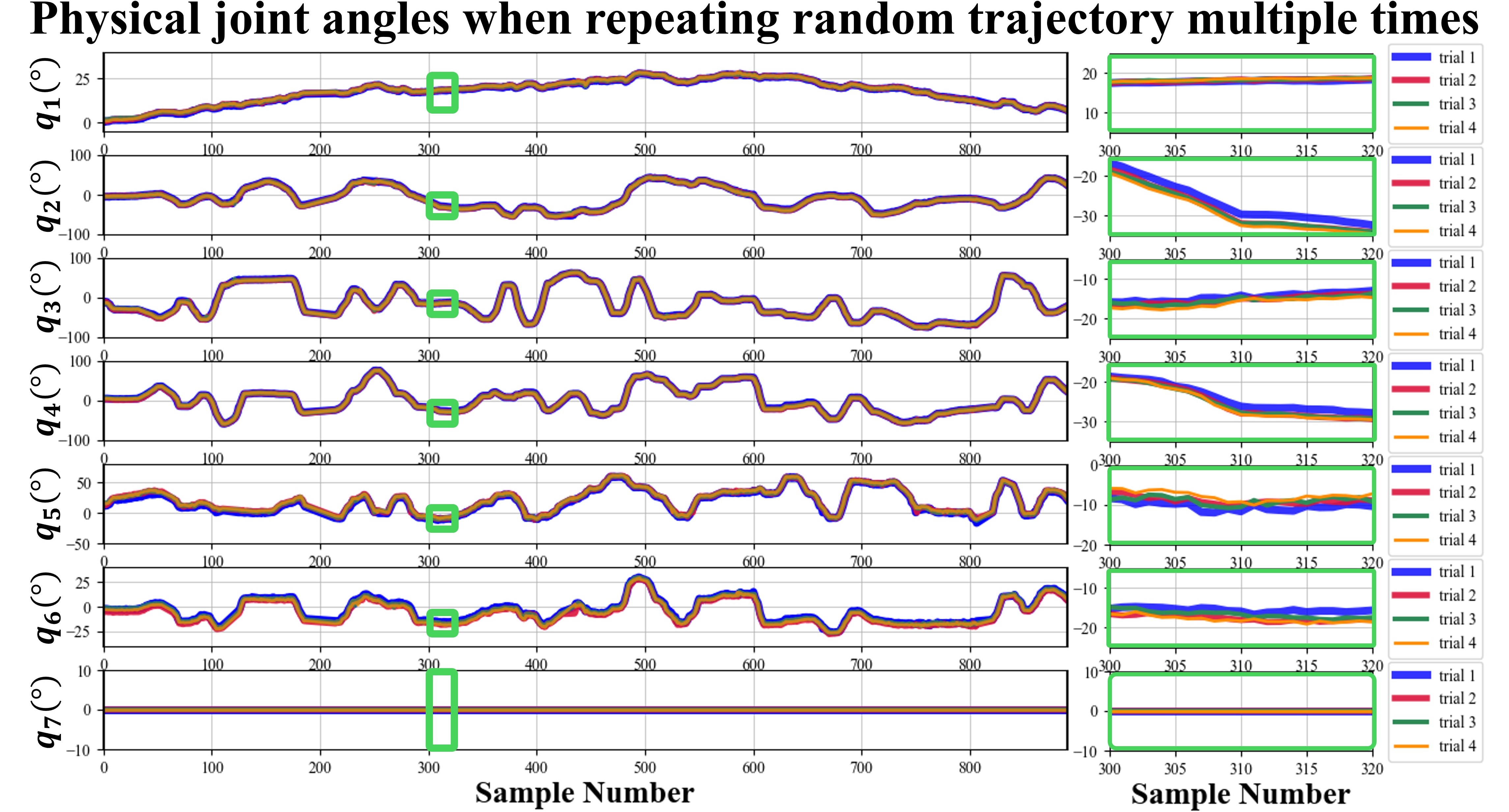}
  \caption{\textbf{Physical Joint Angles during Repeated Random Trajectory Trials}: This figure shows the physical joint angles for a random trajectory consisting of 1000 commanded joint angles, where all joints are moved simultaneously. The frequency of $q_1$ was deliberately kept lower than the other joints, as translational motion is less frequent compared to pitch and yaw movements. The physical joint angles were recorded across four repetitions of the same commanded random trajectory. As highlighted in the enlarged section, the four trials exhibit minimal differences, indicating strong repeatability.}
  \label{fig:random_trajectory_validation}
\end{figure}

\begin{table}[t!]
\centering
\caption{Average and SD of MAE values across four trials on random trajectories}
\label{tab:random_trajectory_validation}
\resizebox{0.55\linewidth}{!}{%
\begin{tabular}{llllllll} 
\toprule
    & $q_1$ ($^\circ$)    & $q_2$ ($^\circ$)  & $q_3$ ($^\circ$) & $q_4$ ($^\circ$) & $q_5$ ($^\circ$) & $q_6$ ($^\circ$) & $q_7$ ($^\circ$) \\ 
\midrule
AVG & 0.27  & 0.90 & 0.83 & 0.75 & 1.56 & 1.31 & 0.00  \\
SD  & 0.095 & 0.31 & 0.21 & 0.18 & 0.72 & 0.46 & 0.00  \\
\bottomrule
\end{tabular}
}
\end{table}

The hysteresis loops observed for the SAM are depicted in \Cref{hysteresis_loop_validation}. Notably, there is a considerable dead zone, such as in the case of the Segment 1 pitch direction with a translation of 0 mm, where the range spans approximately \( 10^\circ \) to \( 60^\circ \) and \( -60^\circ \) to \( -10^\circ \). These results indicate significant nonlinearities, even within single DOFs. Several factors could contribute to the wide dead zone observed, including 1) friction between the driving cable and joints, 2) elongation of the driving cable, and 3) the material properties of PEEK. Moreover, these hysteresis phenomena could be further amplified when multiple joints are activated simultaneously, inducing coupling effects and adding complexity. For example, in \Cref{hysteresis_line}, even when the \( q_6 \) command is held constant at \( 0^\circ \), the \( \mathbf{q_{phy}} \) of \( q_6 \) continues to change.

To assess the repeatability of the hysteresis, we analyzed the mean absolute errors (MAE) between each loop, as shown in \Cref{loop_repeatability}. As depicted in \Cref{hysteresis_loop_validation}, six loops were plotted for each joint. The commands were executed seven times, and the first loop was excluded as it did not complete the loop formation. The MAE between each pair of loops was calculated, resulting in \( _6C_2 \) MAE values. The average and standard deviation (SD) of the MAE were then computed. As indicated in \Cref{loop_repeatability}, the average difference between loops was approximately \( 0.2^\circ \) for \( q_3 \) and \( q_4 \), and \( 1.3^\circ \) for \( q_5 \). These results demonstrate that the hysteresis exhibits repeatable properties.

Additionally, we validated the repeatability of the system under random trajectories. Random trajectories were generated for \( q_1 \) to \( q_7 \), with 50 random points generated for \( q_3 \) to \( q_5 \), which were interpolated with 20 points via linear interpolation. For \( q_2 \), 25 random points were generated and linearly interpolated through 40 points. For \( q_1 \), 12 random points were generated and linearly interpolated through 80 points. This approach allowed us to adjust the period of each joint, reflecting the fact that in real surgical scenarios, the translation joint values (e.g., \( q_3 \) to \( q_5 \)) are typically not altered as frequently. \( q_6 \) and \( q_7 \) were fixed at \( 0^\circ \). The random trajectories were repeated four times, and the corresponding \( \mathbf{q_{phy}} \) were collected, as shown in \Cref{fig:random_trajectory_validation} and \Cref{tab:random_trajectory_validation}. As illustrated in \Cref{fig:random_trajectory_validation}, the four trials on the random trajectory produced similar results, indicating that the hysteresis is repeatable not only in simple DOF experiments but also in complex, multi-DOF scenarios. The quantitative results in \Cref{tab:random_trajectory_validation} show the average MAE and SD between each trial. With \( _4C_2 \) MAE values, the average MAE ranged from approximately \( 0.27^\circ \) to \( 1.56^\circ \), further confirming the repeatability of the hysteresis even in random trajectories.

\end{document}